\documentclass[Afour,sageh,times]{sagej}

\usepackage[utf8]{inputenc} 
\usepackage[T1]{fontenc}    
\usepackage{url}            
\usepackage{booktabs}       
\usepackage{amsfonts}       
\usepackage{nicefrac}       
\usepackage{microtype}      
\usepackage{amssymb}
\usepackage{subfig}
\usepackage{graphicx,color,subfig}
\usepackage{amsfonts, amssymb, amsmath,verbatim, multirow}
\usepackage{enumerate}
\usepackage{breqn}
\usepackage{amsmath}
\usepackage{algorithm}
\usepackage{algorithmic}
\usepackage{easybmat}
\usepackage{moreverb,url}
\usepackage{ftnright}
\usepackage[toc,page]{appendix}

\usepackage[colorlinks,bookmarksopen,bookmarksnumbered,citecolor=red,urlcolor=red]{hyperref}
\usepackage[framemethod=TikZ]{mdframed}
\DeclareMathOperator*{\argmax}{arg\,max}
\DeclareMathOperator*{\argmin}{arg\,min}
\newcommand{\mb}[1]{{\boldsymbol{#1}}}

\newcommand{\trsp}{{\!\scriptscriptstyle\top}}

\newcommand{\tp}[1]{\text{\tiny#1}}
\newcommand{\ty}[1]{{\scriptscriptstyle{\mathcal{#1}}}}
\newcommand{\diag}{\mathrm{diag}}

\newcommand\scalemath[2]{\scalebox{#1}{\mbox{\ensuremath{\displaystyle #2}}}}

\newcommand\BibTeX{{\rmfamily B\kern-.05em \textsc{i\kern-.025em b}\kern-.08em
T\kern-.1667em\lower.7ex\hbox{E}\kern-.125emX}}

\def\volumeyear{2016}

\begin{document}

\runninghead{Tanwani \textit{et al.}}

\title{Small Variance Asymptotics for Non-Parametric Online Robot Learning}

\author{Ajay Kumar Tanwani\affilnum{1,2} and Sylvain Calinon\affilnum{1}}

\affiliation{\affilnum{1} Idiap Research Institute and EPFL, Switzerland. \\
\affilnum{2} University of California, Berkeley.}

\corrauth{Ajay Kumar Tanwani}

\email{ajay.tanwani@berkeley.edu}

\begin{abstract}
Small variance asymptotics is emerging as a useful technique for inference in large scale Bayesian non-parametric mixture models. This paper analyses the online learning of robot manipulation tasks with Bayesian non-parametric mixture models under small variance asymptotics. The analysis yields a \textit{scalable online sequence clustering} (SOSC) algorithm that is non-parametric in the number of clusters and the subspace dimension of each cluster. SOSC groups the new datapoint in low dimensional subspaces by online inference in a non-parametric \textit{mixture of probabilistic principal component analyzers} (MPPCA) based on Dirichlet process, and captures the state transition and state duration information online in a \textit{hidden semi-Markov model} (HSMM) based on hierarchical Dirichlet process. A task-parameterized formulation of our approach autonomously adapts the model to changing environmental situations during manipulation. We apply the algorithm in a teleoperation setting to recognize the intention of the operator and remotely adjust the movement of the robot using the learned model. The generative model is used to synthesize both time-independent and time-dependent behaviours by relying on the principles of shared and autonomous control. Experiments with the Baxter robot yield parsimonious clusters that adapt online with new demonstrations and assist the operator in performing remote manipulation tasks.
\end{abstract}

\maketitle

\section{Introduction}

A long standing goal in artificial intelligence is to make robots interact with humans in everyday life tasks. Programming by demonstration provides a promising route to bridge this gap. When a set of $T$ datapoints of a manipulation task $\{\mb{\xi}_t\}_{t=1}^T$ with $\mb{\xi}_t\in\mathbb{R}^D$ is provided, it is often useful to encode these observations as a generative model with parameters $\Theta$ (e.g., Gaussian mixture model or hidden Markov model), providing a probability density function $\mathcal{P}(\mb{\xi}_t | \Theta)$. Learning a wide range of tasks requires extracting invariant representations from demonstrations that can generalize in previously unseen situations. Encoding the covariance between the task variables is important to represent movement coordination patterns, synergies, and action-perception couplings. Model selection and scalability of encoding the data in higher dimensional spaces limit the ability of these models to represent these important motor control principles. With the influx of high-dimensional sensory data in robotics, mixture models are useful to compactly encode the data online so that the robots are able to perform under varying environmental situations and across range of different tasks. The goal is to provide an approach to teach new manipulation tasks to robots on-the-fly from a few human demonstrations. Moreover, non-parametric online learning can further be combined with other paradigms such as active learning and/or reinforcement learning for improving the acquired skills.

Adapting statistical learning models online with large scale streaming data is a challenging problem. Bayesian non-parametric treatment of these models provides flexibility in model selection by maintaining an appropriate probability distribution over parameter values, $\mathcal{P}(\mb{\xi}_t) = \int \mathcal{P}(\mb{\xi}_t | \Theta) \mathcal{P}(\Theta) d \Theta$ \citep{Opper99}. Although attractive for encapsulating \emph{a priori} information about the task, the computational overhead of existing sampling-based and variational techniques for inference limit the widespread use of these models.

\begin{figure*}[tbh]
\centering
\includegraphics[trim={3.1cm 5.0cm 3cm 4.3cm},clip,scale = 0.64]{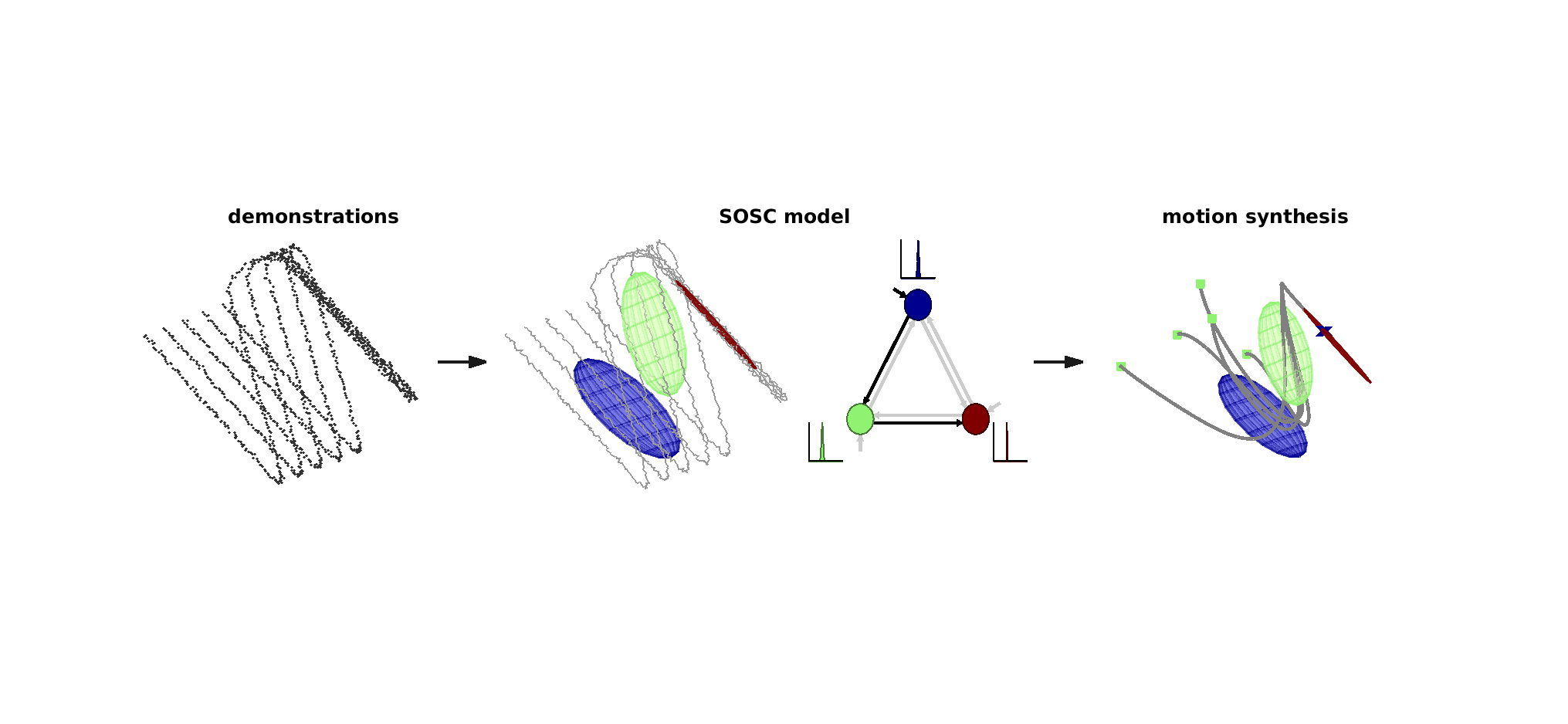} \\
\includegraphics[trim={2.5cm 4.7cm 3.0cm 5.2cm},clip,scale = 0.69]{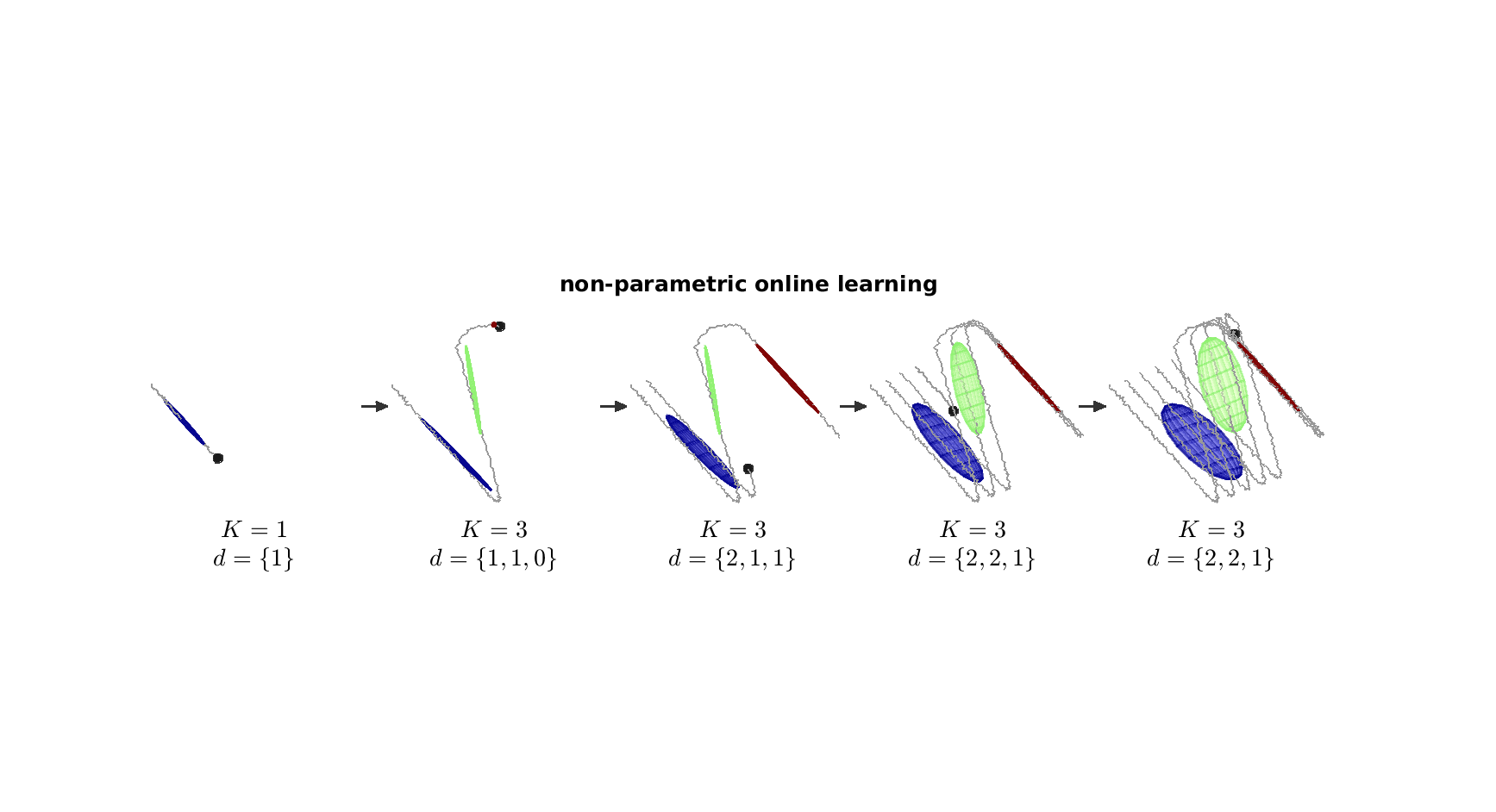} 
\caption{SOSC model illustration with Z-shaped streaming data composed of multiple trajectory samples. The model incrementally clusters the data in its intrinsic subspace. It tracks the transition among states and the state duration steps in a non-parametric manner. The generative model is used to recognize and synthesize motion in performing robot manipulation tasks (see Extension \ref{app: MnE}-$2$).}\label{fig: SOSC_Coceptual}
\end{figure*}

Recent analysis of Bayesian non-parametric mixture models under \textit{small variance asymptotic} (SVA) limit has led to simple deterministic models that scale well with large size applications. For example, as the variances of the mixture model tend to zero in a GMM, the probabilistic model converges to its deterministic counterpart, $k$-means, or to its non-parametric Dirichlet process (DP) version, DP-Means \citep{Kulis12}. SVA analysis of other richer probabilistic models such as dependent DP mixture models \citep{Campbell13}, hierarchical Dirichlet process (HDP) \citep{Jiang12}, infinite latent feature models \citep{Broderick13}, Markov jump processes \citep{Huggins15}, infinite hidden Markov models \citep{Roychowdhury13}, and infinite mixture of probabilistic principal component analysers (MPPCA) \citep{Wang15} leads to similar algorithms that scale well and yet retain the flexibility of non-parametric models. This paper builds upon these advancements to unify online variants of Bayesian non-parametric mixture models under small variance asymptotics for robot learning from demonstrations \citep{Tanwani16b}.

\subsection{Proposed Approach}
We investigate the online learning of robot manipulation tasks under SVA limit of Bayesian non-parametric mixture models. We seek to incrementally update the parameters $\Theta$ with each new observation $\mb{\xi}_{t+1}$ without having to retrain the model in a batch manner and store the demonstration data. We present an online inference algorithm for clustering sequential data, called \textit{scalable online sequence clustering} (SOSC). SOSC incrementally groups the streaming data in low-dimensional subspaces by online inference in the Dirichlet process \textit{mixture of probabilistic principal component analysers} (MPPCA) under small variance asymptotics, while being non-parametric in the number of clusters and the subspace dimension of each cluster. The model tracks the transition between subspaces and the duration of time spent in each subspace by online inference in a \textit{hidden semi-Markov model} (HSMM) based on HDP. A task-parameterized formulation of the SOSC model is used to adapt the model parameters to varying environmental situations in a probabilistic manner \citep{Tanwani2016a}. The proposed approach uses the learning from demonstrations paradigm to teach manipulation tasks to robots in an online and intuitive manner. We show its application in a teleoperation scenario where the SOSC model is built online from the demonstrations provided by the teleoperator to perform remote robot manipulation tasks (see Fig.\ \ref{fig: SOSC_Coceptual} for an overview of our approach).

\subsection{Contributions}

The purpose of this paper is to present an online unsupervised learning framework that is fast and scalable for the encoding of a large range of robot manipulation tasks in a non-parametric manner. The contributions of the paper are:
 
\begin{itemize}

\item Online inference algorithms for DP-GMM, DP-MPPCA and HDP-HSMM under small variance asymptotics,

\item Resulting non-parametric SOSC algorithm for online learning and motion synthesis of high-dimensional robot manipulation tasks,

\item Task-parameterized formulation of the SOSC model to systematically adapt the model parameters to changing situations such as position/orientation/size of the objects,

\item Extension of learning from demonstration to the context of semi-autonomous teleoperation. 


\end{itemize}


%

\textbf{Organization of the paper:} We give a brief overview of unsupervised learning approaches for robot learning in Sec.\ \ref{sec: Background}. Sec.\ \ref{sec: ProblemSetup} formalizes our learning problem of SOSC followed by online inference algorithms of DP-MPPCA and HDP-HSMM in Sec.\ \ref{sec: DP-GMM}, Sec.\ \ref{sec: online_DP_MPPCA} and Sec.\ \ref{sec: online_HDP_HSMM} respectively. In Sec. \ref{sec: SOSC}, we present the overall SOSC algorithm and then evaluate its performance on synthetic data and semi-autonomous teleoperation with the Baxter robot in Sec. \ref{sec: Experiments}. Finally, we conclude the paper with an outlook to our future work.

\section{Background and Related Work} 
\label{sec: Background}

SOSC builds a generative model of the demonstrations online by clustering the streaming data in low-dimensional subspaces and capturing the state transition and state duration information in a non-parametric manner. Incremental online learning poses a unique challenge to the existing robot learning methods with high-dimensional data, model selection, real-time adaptation and adequate accuracy/generalization after observing a fewer number of training samples. An overview of robot learning from demonstration methods can be found in \citep{Schaal03, Argall09, Billard16}.

\textbf{Gaussian mixture models (GMMs)} are widely used to encode local trends in the demonstrations for unsupervised learning problems. The probability density function $\mathcal{P}$ of a GMM with $K$ mixture components is represented as
\begin{equation}
	\mathcal{P}(\mb{\xi}_t | \Theta_{\text{GMM}}) = \sum\limits_{i=1}^K \pi_i\; \mathcal{N}(\mb{\xi}_t|\mb{\mu}_i,\mb{\Sigma}_i), 
	\label{eq:GMM}
\end{equation} 
where $\mathcal{N}(\mb{\mu}_i,\mb{\Sigma}_i)$ is the multivariate Gaussian distribution with parameters $\Theta_{\text{GMM}}$ containing prior $\pi_i \in \mathbb{R}$, mean $\mb{\mu}_i \in \mathbb{R}^{D}$, and covariance matrix $\mb{\Sigma}_i \in \mathbb{R}^{D \times D}$. \textbf{Hidden Markov models (HMMs)} encapsulate the spatio-temporal information by augmenting a GMM with latent states that sequentially evolve over time in the demonstrations. The parameter set now additionally contains the transition matrix and the initial state distribution. HMMs are widely used for time series/sequence analysis in speech recognition, machine translation, DNA sequencing, robotics and many other fields \citep{Rabiner89}. HMMs have been typically used for recognition and generation of movement skills in robotics \citep{Asfour08,Calinon10,Lee10,Vakanski12}. A number of variants of HMMs have been proposed to address some of its shortcomings, including: 1) how to bias learning towards models with longer self-dwelling states, 2) how to robustly estimate the parameters with high-dimensional noisy data, 3) how to adapt the model with newly observed data, and 4) how to estimate the number of states that the model should possess. 



Variants based on \textbf{Hidden semi-Markov models (HSMMs)} replace the self-transition probabilities of staying in a state with an explicit model of state duration \citep{Yu10}. This helps the generative system to adequately represent movements and behaviors with longer state dwell times for learning robot manipulation tasks \citep{Tanwani2016a}.

\textbf{Subspace clustering} methods perform segmentation and dimensionality reduction simultaneously to encode human demonstrations with piecewise planar segments \citep{Schaal07}. Statistical subspace clustering methods impose a parsimonious structure on the covariance matrix to reduce the number of parameters that can be robustly estimated \citep{Bouveyron13}. For example, a \textit{mixture of factor analyzers} (MFA) performs subspace clustering by assuming the structure of the covariance to be of the form \citep{McLachlan03}
\begin{equation}\label{Eq: MFA1}
  \mb{\Sigma}_i = \mb{\Lambda}_i^{d}\mb{\Lambda}_i^{d^\trsp}+\mb{\Psi}_i,
\end{equation}
where $\mb{\Lambda}_i^{d}\in\mathbb{R}^{D\times d}$ is the \emph{factor loadings matrix} with $d\!<\!D$ for parsimonious representation of the data, and $\mb{\Psi}_i\in\mathbb{R}^{D\times D}$ is the diagonal noise matrix. Other extensions such as sharing the parameters of the covariance in a semi-tied manner aligns the mixture components for robust encoding of the demonstrations \citep{Tanwani2016a}. 

\textbf{Online/Incremental learning} methods update the model parameters with streaming data, without the need to re-train the model in a batch manner \citep{Neal99, Song05}. Non-parametric regression methods have been commonly used in this context such as locally weighted projection regression \citep{Vijayakumar05}, sparse online Gaussian process regression \citep{Gijsberts13} and their fusion with local Gaussian process regression \citep{Nguyen15}, see \cite{Stulp15} for a review. \cite{Kulic08} used HMMs to incrementally group whole-body motions based on their relative distance in HMM space. \cite{Lee10IROS} presented an iterative motion primitive refinement approach with HMMs. \cite{Kronander15} locally reshaped an existing dynamical system with new demonstrations in an incremental manner while preserving its stability. \cite{Hoyos16} experimented with different strategies to incrementally add demonstrations to a task-parametrized GMM. \cite{Bruno2016} learned autonomous behaviours for a flexible surgical robot by online clustering with DP-means. 

\textbf{Bayesian non-parametric} treatment of HMMs/HSMMs automates the number of states selection procedure by Bayesian inference in a model with infinite number of states \citep{Beal02,Johnson13}. \cite{Niekum12} used the Beta Process Autoregressive HMM for learning from unstructured demonstrations. \cite{Krishnan15} defined a hierarchical non-parametric Bayesian model to identify the transition structure between states with a linear dynamical system. Figueroa et al. used the transformation invariant covariance matrix for encoding tasks with a Bayesian non-parametric HMM \cite{Figueroa17}. Inferring the maximum \emph{a posteriori} distribution of the parameters in non-parametric models, however, is often difficult. Markov Chain Monte Carlo (MCMC) sampling or variational methods are required, which are difficult to implement and often do not scale with the size of the data. Small variance asymptotic analysis of these methods provide a trade-off by yielding simple and scalable hard clustering non-parametric algorithms \citep{Kulis12}. Other prominent applications of SVA include feature learning \citep{Broderick13}, dimensionality reduction \citep{Wang15} and time-series analysis \citep{Roychowdhury13}.

This paper builds upon these advancements in small variance asymptotic analysis of Bayesian non-parametric mixture models. We present a non-parametric online unsupervised framework for robot learning from demonstrations, which scales well with sequential high-dimensional data. We formulate online inference algorithms of HDP-HSMM and DP-MPPCA under small variance asymptotics in Sec. \ref{sec: online_HDP_HSMM} and Sec. \ref{sec: online_DP_MPPCA} respectively. We then present a task-parameterized generative model online for encoding and motion synthesis of robot manipulation tasks in Sec. \ref{sec: SOSC}.


\section{Problem Setup} \label{sec: ProblemSetup}

Let us consider the streaming observation sequence $\{\mb{\xi}_1, \ldots, \mb{\xi}_t \}$ with $\mb{\xi}_t\in\mathbb{R}^D$ obtained at current time step $t$ while demonstrating a manipulation task. The corresponding hidden state sequence $\{z_1, \ldots, z_t\}$ with $z_t\in \{1, \ldots, K\}$ belongs to the discrete set of $K$ cluster indices at time $t$, and the observation $\mb{\xi}_t$ is drawn from a multivariate Gaussian with mixture coefficients $\pi_{t,i} \in \mathbb{R}$, mean $\mb{\mu}_{t,i} \in \mathbb{R}^{D}$ and covariance $\mb{\Sigma}_{t,i} \in \mathbb{R}^{D \times D}$ at time $t$. 
\begin{figure}[!tb]
\centering
\includegraphics[trim={13.8cm 1.7cm 10.0cm 0.75cm},clip,scale = 0.56]{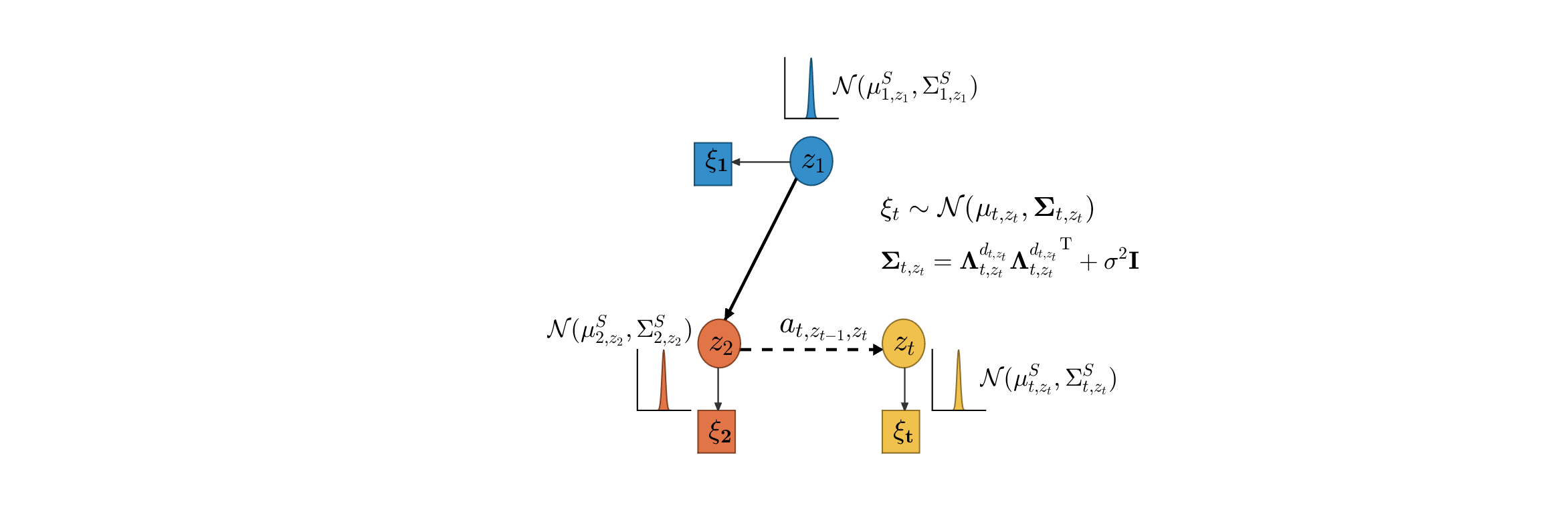}
\caption{SOSC representation using non-parametric HSMM with MPPCA as observation distribution given the streaming data $\mb{\xi}_1, \mb{\xi}_2, \ldots \mb{\xi}_t$.}\label{fig: SOSC_GM}
\end{figure}

We seek to update the parameters online upon observation of a new datapoint $\mb{\xi}_{t+1}$, such that the datapoint can be discarded afterwards. Small variance asymptotic (SVA) analysis implies that the covariance matrix $\mb{\Sigma}_{t,i}$ of all the Gaussians reduces to the isotropic noise $\sigma^{2}$, i.e., $\mb{\Sigma}_{t,i} \approx \lim_{\sigma^{2} \rightarrow 0} \sigma^{2} \mb{I}$ \citep{Kulis12, Broderick13, Roychowdhury13}. Note that if the covariance matrices $\mb{\Sigma}_{t,i}$ of all the mixture components in a GMM are set equal to the isotropic matrix $\sigma^{2} \mb{I}$, the expected value of the complete log-likelihood of the data a.k.a.\ the auxiliary function, $\mathcal{Q}(\Theta_{\text{GMM}},\Theta_{\text{GMM}}^{\; \text{old}}) =  \mathbb{E} \left \{ \log \mathcal{P}(\mb{\xi}_t, z_t|\Theta_{\text{GMM}}) \; \big | \; \mb{\xi}_t, \Theta_{\text{GMM}}^{\;  \text{old}} \right \}$, takes the form \citep{Dempster77}
\begin{equation}
\sum_{i=1}^{K} \mathcal{P}(i|\mb{\xi}_t, \Theta_{\text{GMM}}^{\; \text{old}})\left( \log \pi_{t,i} - \frac{D}{2} \log 2\pi \sigma^{2} - \frac{\|\mb{\xi}_t - \mb{\mu}_{t,i}\|_2^{2}}{2\sigma^{2}} \right).
\end{equation} 
Applying the small variance asymptotic limit to the auxiliary function with $\lim_{\sigma^{2} \rightarrow 0} \mathcal{Q}(\Theta_{\text{GMM}},\Theta_{\text{GMM}}^{\; \text{old}})$, the last term $\frac{\|\mb{\xi}_t - \mb{\mu}_{t,i}\|_2^{2}}{2\sigma^{2}}$ dominates the objective function and the maximum likelihood estimate reduces to the $k$-means problem,\footnote{SVA analysis of the Bayesian non-parametric GMM leads to the DP-means algorithm \cite{Kulis12}. Similarly, SVA analysis of the HMM yields the segmental $k$-means problem \cite{Roychowdhury13}.} i.e., 
\begin{equation}
\max \mathcal{Q}(\Theta_{\text{GMM}},\Theta_{\text{GMM}}^{\; \text{old}}) = \argmin_{z_t, \mb{\mu}_t} \| \mb{\xi}_t - \mb{\mu}_{t, z_t} \|_2^{2}.
\end{equation} 

By restricting the covariance matrix to an isotropic/spherical noise, the number of parameters grows up to a constant with the dimension of datapoint $D$. Although attractive for scalability and parsimonious structure, such decoupling cannot encode the important motor control principles of coordination, synergies and action-perception couplings. Consequently, we further assume that the $i^{th}$ output Gaussian groups the observation $\mb{\xi}_t$ in its intrinsic low-dimensional affine subspace of dimension $d_{t,i}$ at time $t$ with projection matrix $\mb{\Lambda}_{t,i}^{d_{t,i}} \in \mathbb{R}^{D \times d_{t,i} }$, such that $d_{t,i} < D$ and $\mb{\Sigma}_{t,i} = \mb{\Lambda}_{t,i}^{d_{t,i}} \mb{\Lambda}_{t,i}^{{d_{t,i}}^{\trsp}} + \sigma^{2} \mb{I}$. Under this assumption, we apply the small variance asymptotic limit on the remaining $(D - d_{t,i})$ dimensions to encode the most important coordination patterns while being parsimonious in the number of parameters.

In order to encode the temporal information among the mixture components, let $\mb{a}_t \in \mathbb{R}^{K \times K}$ with $a_{t, i, j} \triangleq P(z_t = j | z_{t-1} = i)$ denote the transition probability of moving from state $i$ at time $t-1$ to state $j$ at time $t$. The parameters $\{\mu_{t, i}^{S}, \Sigma_{t,i}^{S}\}$ represent the mean and the standard deviation of staying $s$ consecutive time steps in state $i$ estimated by a Gaussian $\mathcal{N}(s|\mu_{t,i}^{S},\Sigma_{t,i}^{S})$. The hidden state follows a multinomial distribution with $z_t \sim \text{Mult}(\mb{\pi}_{z_{t-1}})$ where $\mb{\pi}_{z_{t-1}} \in \mathbb{R}^{K}$ is the next state transition distribution over state ${z_{t-1}}$, and the observation $\mb{\xi}_t$ is drawn from the output distribution of state $j$, described by a multivariate Gaussian with parameters $\{\mb{\mu}_{t,j}, \mb{\Sigma}_{t,j}\}$ (see Fig.\ \ref{fig: SOSC_GM} for graphical representation of the SOSC problem). The $K$ Gaussian components constitute a GMM augmented with the state transition and the state duration model to capture the sequential pattern in the demonstrations (see Appx.\ \ref{app: symbols} for the notations used in the paper). 

The overall parameter set of SOSC is represented by $\Theta_{t,\text{SOSC}} = \Big\{\mb{\mu}_{t,i},\mb{\Sigma}_{t,i}, \{a_{t, i,m} \}_{m=1}^{K}, \mu_{t, i}^{S},\Sigma_{t, i}^{S} \Big\}_{i=1}^{K}$.\footnote{With a slight abuse of notation, we represent the parameters with an added subscript $t$ for online learning. For example, $\Theta_{t,h}$ denotes the parameters of $\Theta_h$ at time $t$.} We are interested in updating the parameter set $\Theta_{t,\text{SOSC}}$ online upon observation of a new datapoint $\mb{\xi}_{t+1}$, such that the datapoint can be discarded afterwards. We first apply the Bayesian non-parametric treatment to the underlying mixture models and formulate online inference algorithms for DP-GMM, DP-MPPCA and HDP-HSMM under small variance asymptotics in Sec. \ref{sec: DP-GMM}, \ref{sec: online_DP_MPPCA}, and \ref{sec: online_HDP_HSMM}. This results in a non-parametric online approach to robot learning from demonstrations presented in Sec. \ref{sec: SOSC}.

\section{SVA of DP-GMM}\label{sec: DP-GMM}
In this section, we review the fundamentals of Bayesian non-parametric extension of GMM under small variance asymptotics using the parameter subset $\Theta_{\text{GMM}}= \{\pi_i, \mb{\mu}_i, \mb{\Sigma}_i\}_{i=1}^{K}$ and present a simple approach for online update of the parameters. 

\subsection{Dirichlet Process GMM (DP-GMM)}
Consider a Bayesian non-parametric GMM with \textit{Chinese Restaurant Process} (CRP) prior over the cluster assignment with $\alpha$ as concentration parameter, $z_t \sim \text{CRP}(\alpha)$, and non-informative prior over cluster means with $\varrho^{2}$ as small constant, $\mb{\mu}_{i} \sim \mathcal{N}(\mb{0}, \varrho^{2}\mb{I}_D)$. The likelihood function for a set of datapoints is evaluated as
\begin{equation}
\mathcal{P}(\mb{\xi}_t | \mb{z}, \mb{\mu}) = \prod_{i=1}^{K} \prod_{t=1}^{T} \mathcal{N}(\mb{\xi}_t | \mb{\mu}_i, \sigma^{2}\mb{I}).
\end{equation}
The parameters $\mb{z}$ and $\mb{\mu}$ are obtained by maximizing the posterior distribution
\begin{equation}
\argmax_{K, \mb{z}, \mb{u}} \mathcal{P}(\mb{z}, \mb{\mu} | \mb{\xi}_t) \propto \argmin_{K, \mb{z}, \mb{u}} - \log \mathcal{P}(\mb{\xi}_t, \mb{z}, \mb{\mu}).
\end{equation} 
Computing the joint posterior distribution and setting $\alpha = \exp(-\frac{\lambda}{2\sigma^{2}})$
\begin{multline}
\mathcal{P}(\mb{\xi}_t, \mb{z}, \mb{\mu}) = \mathcal{P}(\mb{\xi}_t | \mb{z}, \mb{\mu}) \; \mathcal{P}(\mb{z}) \; \mathcal{P}(\mb{\mu}) \\
= \prod_{i=1}^{K} \prod_{t=1}^{T} \mathcal{N}(\mb{\xi}_t | \mb{\mu}_i, \sigma^{2}\mb{I}) \; \text{CRP}(\exp(-\frac{\lambda}{2\sigma^{2}})) \; \mathcal{N}(\mb{0}, \varrho^{2}\mb{I}_D).
\end{multline}
Taking the $\log$ of the joint posterior distribution and applying the SVA limit $\lim_{\sigma^{2} \rightarrow 0}$ yields the DP-means algorithm \citep{Kulis12}. The limit pushes the posterior mass on one of the clusters leading to a deterministic assignment based on the distance of the datapoint to the nearest cluster. The resulting loss function $\mathcal{L}(\mb{z},\mb{\mu})$ to optimize is given as
\begin{multline}
\argmin_{K, \mb{z}, \mb{u}} \lim_{\sigma^{2} \rightarrow 0} -\log \mathcal{P}(\mb{\xi}_t, \mb{z}, \mb{\mu}) \approx \argmin_{K, \mb{z}, \mb{u}} \mathcal{L}(\mb{z},\mb{\mu}) \\ =  \argmin_{K, \mb{z}, \mb{u}} \sum_{i=1}^{K}\sum_{t=1}^{T} \| \mb{\xi}_{t} - \mb{\mu}_i \|_2^{2} \; + \; \lambda K.
\end{multline}
The algorithm is similar to $k$-means algorithm except that it is non-parametric in the number of clusters. The algorithm iteratively assigns the datapoint(s) to its nearest cluster center, and if any of the datapoints are farther away from the cluster center than a certain threshold $\lambda$, a new cluster is created with the distant datapoints and a penalty $\lambda$ added to the loss function. The algorithm converges to a local minimum just like the $k$-means algorithm. 

\subsection{Online Inference in DP-GMM}
In the online setting, we want to update the parameters $\Theta_{t,\text{GMM}}$ with each new observation $\mb{\xi}_{t+1}$. The update consists of the cluster assignment step and incremental update of parameters step.

\subsubsection{Cluster Assignment $z_{t+1}$:} 
In the online setting, the cluster assignment $z_{t+1}$ for new datapoint $\mb{\xi}_{t+1}$ is based on the distance of the datapoint to the existing cluster means. If the minimum distance is greater than a certain threshold $\lambda$, a new cluster is initialized with that datapoint; otherwise the assigned cluster prior, mean and the corresponding number of datapoints $w_{t+1,z_{t+1}}$ are incrementally updated. We can thus write, 
\begin{equation}
z_{t+1} = \argmin_{j=1:K+1}\begin{cases} \label{Eq: DP-Means1}
		\| \mb{\xi}_{t+1} - \mb{\mu}_{t,j} \|_2^{2} ,& \text{if } j \leq K\\
	\qquad \lambda,              & \text{otherwise.}
	\end{cases}
\end{equation} 

\subsubsection{Parameters Update $\Theta_{t+1,\text{GMM}}$:} 
Given the cluster assignment $z_{t+1} = i$ and the covariance matrix set to $\mb{\Sigma}_{t,i} = \sigma^{2}\mb{I}$, the parameters are updated with 
\begin{align}
\pi_{t+1,i} &= \frac{1}{t+1}\Big(t \pi_{t,i} + 1\Big),\nonumber \\ \mb{\mu}_{t+1,i} &= \frac{1}{w_{t,i} +1}\Big(w_{t,i} \mb{\mu}_{t,i} + \mb{\xi}_{t+1}\Big), \label{Eq: Dp-means2} 
\end{align}
where $w_{t,i}$ is the weight assigned to the $i$-th cluster parameter set at time $t$ to control the effect of the parameter update with the new datapoint at time $t+1$ relative to the updates seen till time $t$ (see next section for updates of $w_{t+1,i}$). 

\textbf{Loss function $\mathcal{L}(z_{t+1}, \mb{\mu}_{t+1,z_{t+1}})$}: The loss function optimized at time step $t+1$ is given as
\begin{multline}
	\mathcal{L}(z_{t+1}, \mb{\mu}_{t+1,z_{t+1}}) = 
	\lambda K + \| \mb{\xi}_{t+1} - \mb{\mu}_{t+1,z_{t+1}} \|_2^{2} \\ \; \leq \; \mathcal{L}(z_{t+1},\mb{\mu}_{t, z_{t+1}}).
\end{multline} 
It can be seen that direct application of small variance asymptotic limit with isotropic Gaussians severely limits the model from encoding important coordination patterns/variance in the streaming data. We next apply the limit to discard only the redundant dimensions in a non-parametric manner and project the new datapoint in a latent subspace by online inference in a Dirichlet process mixture of probabilistic principal component analyzers.

\section{Online DP-MPPCA} \label{sec: online_DP_MPPCA}

In this section, we consider the problem formulation with a \textit{mixture of probabilistic principal component analyzers} (MPPCA) using the parameter subset $\Theta_{\text{MPPCA}} = \big\{\mb{\mu}_i,\mb{\Lambda}_i^{d}, d_i \big\}_{i=1}^{K}$. We consider its non-parametric extension with the Dirichlet process under small variance asymptotics and present an algorithm for online inference.

\subsection{Mixture of Probabilistic Principal Component Analyzers (MPPCA)}

The basic idea of MPPCA is to reduce the dimensions of the data while keeping the observed covariance structure. The generative model of MPPCA approximates the datapoint $\mb{\xi}_t$ as a convex combination of $K$ subspace clusters \citep{Tipping99b}
\begin{equation}
	\mathcal{P}(\mb{\xi}_t | \theta_s) = \sum\limits_{i=1}^{K} \mathcal{P}(z_t = i)\;\mathcal{N}(\mb{\xi}_t | \mb{\mu}_i, 
	\mb{\Lambda}_i^{d}\mb{\Lambda}_i^{d^\trsp}+ \sigma_i^2\mb{I}),
\end{equation} 
where $\mathcal{P}(z_t = i)$ is the cluster prior, $\mb{\Lambda}_i^{d}\in\mathbb{R}^{D\times d}$ is the projection matrix with $d < D$ and $d = d_i$, $\sigma_i^2\mb{I}$ is the isotropic noise coefficient for the $i$-th cluster, and the covariance structure is of the form $\mb{\Sigma}_i = \mb{\Lambda}_i^{d}\mb{\Lambda}_i^{d^\trsp}+ \sigma_i^2\mb{I}$.\footnote{Note that MPPCA is closely related to MFA, and uses isotropic noise matrix instead of the diagonal noise matrix used in MFA (see Eq.\ \eqref{Eq: MFA1}).} The model assumes that $\mb{\xi}_t$, conditioned on $z_t = i$, is generated by an affine transformation of $d$-dimensional latent variable $\mb{u}_t \in \mathbb{R}^{d}$ with noise term $\epsilon \in \mathbb{R}^{D}$ such that
\begin{equation}
	\mb{\xi}_t = \mb{\Lambda}_i^{d} \mb{u}_t + \mb{\mu}_i + \mb{\epsilon} , \quad \mb{u}_t \sim \mathcal{N}(\mb{0}, \mb{I}_d), 
	\quad \mb{\epsilon} \sim \mathcal{N}(\mb{0},\sigma_i^2\mb{I}).
\end{equation} 

The model parameters of MPPCA are usually learned using an Expectation-Maximization (EM) procedure \citep{Tipping99b}. But in this case, both the number of clusters $K$ and the subspace dimension of each cluster $d$ need to be specified a priori, which is not always trivial in several domains. 

\subsection{Dirichlet Process MPPCA (DP-MPPCA)}

Bayesian non-parametric extension of MPPCA alleviates the problem of model selection by defining prior distributions over the number of clusters $K$ and the subspace dimension of each cluster $d_i$ \citep{Zhang04,Chen10,Wang15}. Similar to DP-GMM, a CRP prior is placed over the cluster assignment $z_t \sim \text{CRP}(\alpha)$, along with a hierarchical prior over the projection matrix $\mb{\Lambda}_i^{d_i}$ and an exponential prior on the subspace rank $d_i \sim r^{d_i}$ where $r \in (0,1)$. Applying small variance asymptotics on the resulting partially collapsed Gibbs sampler leads to an efficient deterministic algorithm for subspace clustering with an infinite MPPCA \citep{Wang15}. The algorithm iteratively converges by minimizing the loss function
\begin{equation}
	\mathcal{L}(\mb{z}, \mb{d}, \mb{\mu}, \mb{U}) = \lambda K + \lambda_1 \sum_{i=1}^{K}d_i + \sum_{t=1}^{T} 
		\text{dist}(\mb{\xi}_t, \mb{\mu}_{z_t} , \mb{U}_{z_t}^{d})^{2}, 
	\label{Eq: LossF-DP-MPPCA}
\end{equation} 
where $\text{dist}(\mb{\xi}_t, \mb{\mu}_{z_t} , \mb{U}_{z_t}^{d})^{2}$ represents the distance of the datapoint $\mb{\xi}_t$ to the subspace of cluster $z_t$ defined by mean $\mb{\mu}_{z_t}$ and unit eigenvectors of the covariance matrix $\mb{U}_{z_t}^{d}$ (see Eq.\ \eqref{Eq: distSS} below), and $\lambda, \lambda_1$ represent the penalty terms for the number of clusters and the subspace dimension of each cluster respectively. The algorithm optimizes the number of clusters and the subspace dimension of each cluster while minimizing the distance of the datapoints to the respective subspaces of each cluster. Note that the clustering objective is similar to the DP-means algorithm except that the distance to the cluster means is replaced by the distance to the subspace of the cluster and an added penalty is placed on choosing clusters with more subspace dimensions. In other words, DP-GMM is the limiting case of DP-MPPCA with very large penalty on the subspace dimension.

\subsection{Online Inference in DP-MPPCA}

In the online setting, we seek to incrementally update the parameters $\Theta_{t,\text{MPPCA}}$ ($\Theta_{\text{MPPCA}}$ at time $t$) with the new observation $\mb{\xi}_{t+1}$ without having to retrain the model in a batch manner and store the demonstration data. The parameters are updated in an online manner in two steps: the cluster assignment step followed by the parameter updates step.

\subsubsection{Cluster Assignment $z_{t+1}$:} 
The cluster assignment $z_{t+1}$ of $\mb{\xi}_{t+1}$ in the online case follows the same principle as in Eq.\ \eqref{Eq: DP-Means1}, except the distance is now computed from the subspace of a cluster $\text{dist}(\mb{\xi}_{t+1}, \mb{\mu}_{t,i} , \mb{U}_{t,i}^{d_{t,i}})^{2}$, defined using the difference between the mean-centered datapoint and the mean-centered datapoint projected upon the subspace $\mb{U}_{t,i}^{d_{t,i}} \in \mathbb{R}^{D \times d_{t,i}}$ spanned by the $d_{t,i}$ unit eigenvectors of the covariance matrix, i.e.,
\begin{multline}
		\text{dist}(\mb{\xi}_{t+1}, \mb{\mu}_{t,i} , \mb{U}_{t,i}^{d_{t,i}}) = \\
	{\Big\|(\mb{\xi}_{t+1} - \mb{\mu}_{t,i}) - 
	\rho_i \mb{U}_{t,i}^{d_{t,i}}\mb{U}_{t,i}^{{d_{t,i}}^\trsp} (\mb{\xi}_{t+1} - \mb{\mu}_{t,i}) \Big\|}_2, 
	\label{Eq: distSS}
\end{multline} 
where
\begin{equation*}
	\rho_i = \exp\left(-\frac{\|\mb{\xi}_{t+1} - \mb{\mu}_{t,i}\|_2^{2}}{ b_m}\right)
\end{equation*}
weighs the projected mean-centered datapoint according to the distance of the datapoint from the cluster center $(0 < \rho_i \leq 1)$. Its effect is controlled by the bandwidth parameter $b_m$. If $b_m$ is large, then the far away clusters have a greater influence; otherwise nearby clusters are favored. Note that $\rho_j$ assigns more weight to the projected mean-centered datapoint for the nearby clusters than the distant clusters to limit the size of the cluster/subspace. Note that our subspace distance formulation is different from \citep{Wang15} as we weigh the subspace of the nearby clusters more than the distant clusters. This allows us to avoid clustering all the datapoints in the same subspace (near or far) together. The cluster assignment is deterministically updated using
\begin{equation}
	z_{t+1} = \argmin_{i = 1:K + 1} \begin{cases} \label{Eq: DP-PCA-4}
		\text{dist}(\mb{\xi}_{t+1}, \mb{\mu}_{t,i} , \mb{U}_{t,i}^{d_{t,i}})^{2},& \text{if } i \leq K\\
	\lambda,              & \text{otherwise.}
	\end{cases}
\end{equation} 

\subsubsection{Parameter Updates $\Theta_{t+1,\text{MPPCA}}$:} 
Given the cluster assignment $z_{t+1} = i$ at time $t+1$, the prior and mean of the assigned cluster are updated in the same way as DP-GMM (see Eq.\ \eqref{Eq: Dp-means2}). Depending upon the nature of the streaming data, $w_{t+1,i}$ can be updated as follows\footnote{Note that when the transition dynamics and the observations are modeled with a linear Gaussian model, the parameter updates for the mean and the covariance can be updated in closed form with the use of a Kalman filter.}:
\begin{itemize}
\item For stationary online learning problems where the data is sampled from some fixed distribution, we update the weight $w_{t+1,i}$ linearly with the number of instances belonging to that cluster, namely 
\begin{equation} \label{Eq: DP-PCA-StationaryWeight}
	w_{t+1,i} = w_{t,i} + 1, \qquad w_{0,i} = 1.
\end{equation}
\item For non-stationary online learning problems where the distribution of streaming data varies over time, we update the weight vector based on the \textit{eligibility trace} that takes into account the temporary occurrence of visiting a particular cluster.\footnote{Eligibility traces are commonly used in reinforcement learning to evaluate the state for undergoing learning changes in temporal-difference learning \citep{Sutton98}.} The trace indicates how much a cluster is eligible for undergoing changes with the new parameter update. The trace is updated such that the weights of all the clusters are decreased by the discount factor $\zeta \in (0,1)$ and the weight of the visited cluster is incremented, i.e., the more often a state is visited, the higher is the eligibility weight of all the previous updates relative to the new parameter update, namely
\begin{equation}
	w_{t+1,i} = \begin{cases} 
	\zeta w_{t,i} + 1,& \text{if } i = z_{t+1}\\
	\zeta w_{t,i},    & \text{if } i \neq z_{t+1}.
	\end{cases}
	\label{Eq: DP-PCA-ELTR}
\end{equation}
\item For non-stationary problems where learning is continuous and may not depend upon the number of datapoints, the weight vector is kept constant $w_{t+1,i} = w_{t,i} = w^{*}$ at all time steps as a step-size parameter.
\end{itemize} 

The covariance matrix could then be updated online as
\begin{multline}
	\mb{\bar{\Sigma}}_{t+1,i} = \frac{w_{t,i}}{w_{t,i} +1} \; \mb{\Sigma}_{t,i} + \\ 
	\frac{w_{t,i}}{(w_{t,i} +1)^2} \; (\mb{\xi}_{t+1} - \mb{\mu}_{t+1,i})(\mb{\xi}_{t+1} - \mb{\mu}_{t+1,i})^{\trsp}. 
	\label{Eq: OnlineDP-PCA Sigma}
\end{multline} 
However, updating the covariance matrix online in $D$-dimensional space can be prohibitively expensive for even moderate size problems. To update the covariance matrix in its intrinsic lower dimension, similarly to \citep{Bellas2013}, we compute $\mb{g}_{t+1,i}\in\mathbb{R}^{d_i}$ as the projection of datapoint $\mb{\xi}_{t+1}$ onto the existing set of basis vectors of $\mb{U}_{t,i}^{d_{t,i}}$. Note that the cardinality of basis vectors is different for each covariance matrix. If the datapoint belongs to the subspace of $\mb{U}_{t,i}^{d_{t,i}}$, the retro-projection of the datapoint in its original space, as given by the residual vector $\mb{p}_{t+1,i}\in\mathbb{R}^{D}$, would be a zero vector; otherwise the residual vector belongs to the null space of $\mb{U}_{t,i}^{d_{t,i}}$, and its unit vector $\tilde{\mb{p}}_{t+1,i}$ needs to be added to the existing set of basis vectors, i.e.,
\begin{align}
	\mb{g}_{t+1,i} &= {\mb{U}_{t,i}^{d_{t,i}}}^{\trsp}(\mb{\xi}_{t+1} - \mb{\mu}_{t,i}), \nonumber\\ 
	\mb{p}_{t+1,i} &= (\mb{\xi}_{t+1} - \mb{\mu}_{t,i}) - \mb{U}_{t,i}^{d_{t,i}} \; \mb{g}_{t+1,i}, \nonumber\\
	\tilde{\mb{p}}_{t+1,i} &= \begin{cases}
	\frac{\mb{p}_{t+1,i}}{{\| \mb{p}_{t+1,i} \|}_2},& \text{if } {\| \mb{p}_{t+1,i} \|}_2 > 0\\
		\mb{0}_D,              & \text{otherwise.}
	\end{cases} \nonumber
\end{align} 

The new set of basis vectors augmented with the unit residual vector is represented as
\begin{equation}
\mb{U}_{t+1,i}^{d_{t,i}} = [\mb{U}_{t,i}^{d_{t,i}} \ , \ \tilde{\mb{p}}_{t+1,i}] \ \mb{R}_{t+1,i},  
	\label{Eq: online DP-PCA} 
\end{equation} 
where $\mb{R}_{t+1,i}\in\mathbb{R}^{(d_{t,i}+1) \times (d_{t,i}+1)}$ is the rotation matrix to incrementally update the augmented basis vectors. $\mb{R}_{t+1,i}$ is obtained by simplifying the eigendecomposition problem
\begin{equation}
	\mb{\bar{\Sigma}}_{t+1,i}  = \mb{U}_{t+1,i}^{d_{t,i}} \; \mb{\Sigma}_{t+1,i}^{(\diag)} \; {\mb{U}_{t+1,i}^{d_{t,i}}}^{\trsp}.
\end{equation} 

Substituting the value of $\mb{\bar{\Sigma}}_{t+1,i}$ from Eq.\ \eqref{Eq: OnlineDP-PCA Sigma} and $\mb{U}_{t+1,i}^{d_{t,i}}$ from \eqref{Eq: online DP-PCA} yields the reduced eigendecomposition problem of size $(d_{t,i}+1) \times (d_{t,i}+1)$ with
\begin{multline}
	\frac{w_{t,i}}{w_{t,i} +1} \scalemath{0.99}{\begin{bmatrix} \mb{\Sigma}_{t,i}^{(\diag)} & \mb{0}_{d_{t,i}} \\ 
	\mb{0}_{d_{t,i}}^{\trsp} & 0\end{bmatrix}} 
	+  \frac{w_{t,i}}{(w_{t,i} +1)^2} \\
	\scalemath{0.99}{\begin{bmatrix}  \mb{g}_{t+1,i} \; \mb{g}_{t+1,i}^{\trsp} & \nu_i\mb{g}_{t+1,i} \\ 
	\nu_i\mb{g}_{t+1,i}^{\trsp}	& \nu_i^{2}\end{bmatrix}}  
	= \mb{R}_{t+1,i} \; \mb{\Sigma}_{t+1,i}^{(\diag)} \; \mb{R}_{t+1,i}^{\trsp},
\end{multline} 
where $\nu_i = \tilde{\mb{p}}_{t+1,i}^{\trsp}(\mb{\xi}_{t+1} - \mb{\mu}_{t+1,i})$. Solving for $\mb{R}_{t+1,i}$ and substituting it in Eq.\ \eqref{Eq: online DP-PCA} gives the required updates of the basis vectors in a computationally and memory efficient manner. The subspace dimension of the $i$-th mixture component is updated by keeping an estimate of the average distance vector $\mb{\bar{e}}_{t,i} \in \mathbb{R}^{D}$ whose $k$-th element represents the mean distance of the datapoints to the $(k-1)$ subspace basis vectors of $\mb{U}_{t,i}^{k}$ for the $i$-th cluster. Let us denote $\mb{\delta}_i$ as the vector measuring the distance of the datapoint $\mb{\xi}_{t+1}$ to each of the subspaces of $\mb{U}_{t,i}^{k}$ for the $i$-th cluster where $k = \{0 \ldots (d_{t,i} + 1)\}$, i.e., 
\begin{equation}
	\mb{\delta}_i = 
	\begin{bmatrix} \text{dist}(\mb{\xi}_{t+1}, \mb{\mu}_{t+1,i} , \mb{U}_{t+1,i}^{0})^{2} \\ \vdots \\ 
	\text{dist}(\mb{\xi}_{t+1}, \mb{\mu}_{t+1,i} , \mb{U}_{t+1,i}^{d_{t,i} + 1 })^{2} \end{bmatrix},
\end{equation} 
where $\text{dist}(\mb{\xi}_{t+1}, \mb{\mu}_{t+1,i} , \mb{U}_{t+1,i}^{0})^{2}$ is the distance to the cluster subspace with $0$ dimension (the cluster center point), $\text{dist}(\mb{\xi}_{t+1}, \mb{\mu}_{t+1,i} , \mb{U}_{t+1,i}^{1})^{2}$ is the distance to the cluster subspace with $1$ dimension (the line), and so on. The average distance vector $\mb{\bar{e}}_{t+1,i}$ and the subspace dimension $d_{t+1,i}$ are incrementally updated as
\begin{align}
	\mb{\bar{e}}_{t+1,i} &= \frac{1}{w_{t,i} + 1} \Big(w_{t,i} \mb{\bar{e}}_{t,i} + \mb{\delta}_{i}\Big), \\ 
	d_{t+1,i} &= \argmin_{d = 0:D-1} \Big\{ \lambda_1 d  + \mb{\bar{e}}_{t+1,i} \Big\}.  
	\label{Eq: OnlineDPPCA-di}
\end{align} 

Given the updated set of basis vectors, the projection matrix and the covariance matrix are updated as
\begin{gather}
	\mb{\Lambda}_{t+1,i}^{d_{t+1,i}} = \mb{U}_{t+1,i}^{d_{t+1,i}} \; \sqrt{\mb{\Sigma}_{t+1,i}^{(\diag)}}, \\ 
	\mb{\Sigma}_{t+1,i} = \mb{\Lambda}_{t+1,i}^{d_{t+1,i}} \; {\mb{\Lambda}_{t+1,i}^{d_{t+1,i}}}^{\trsp} + \sigma^{2}\mb{I}. 
	\label{Eq: OnlineDPPCA-SigmaU}
\end{gather} 
\begin{figure}[!tb]
\centering
\includegraphics[trim={1.6cm 3.7cm 2.0cm 4.2cm},clip,scale = 0.64]{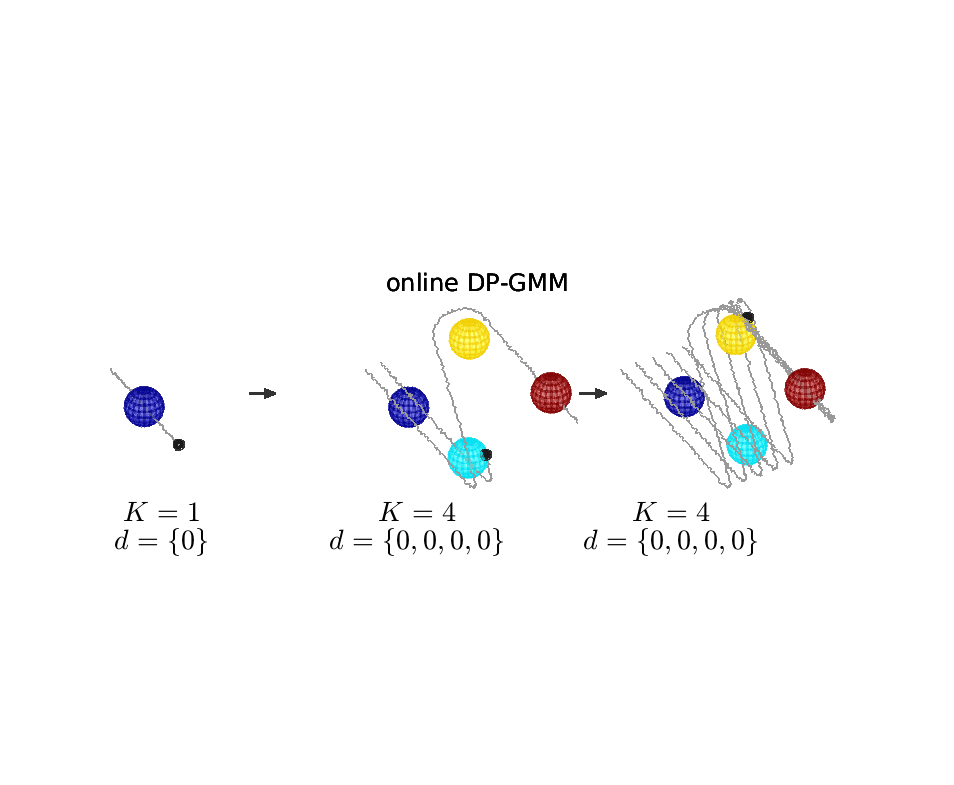} \\
\includegraphics[trim={1.6cm 3.7cm 2.0cm 4.2cm},clip,scale = 0.64]{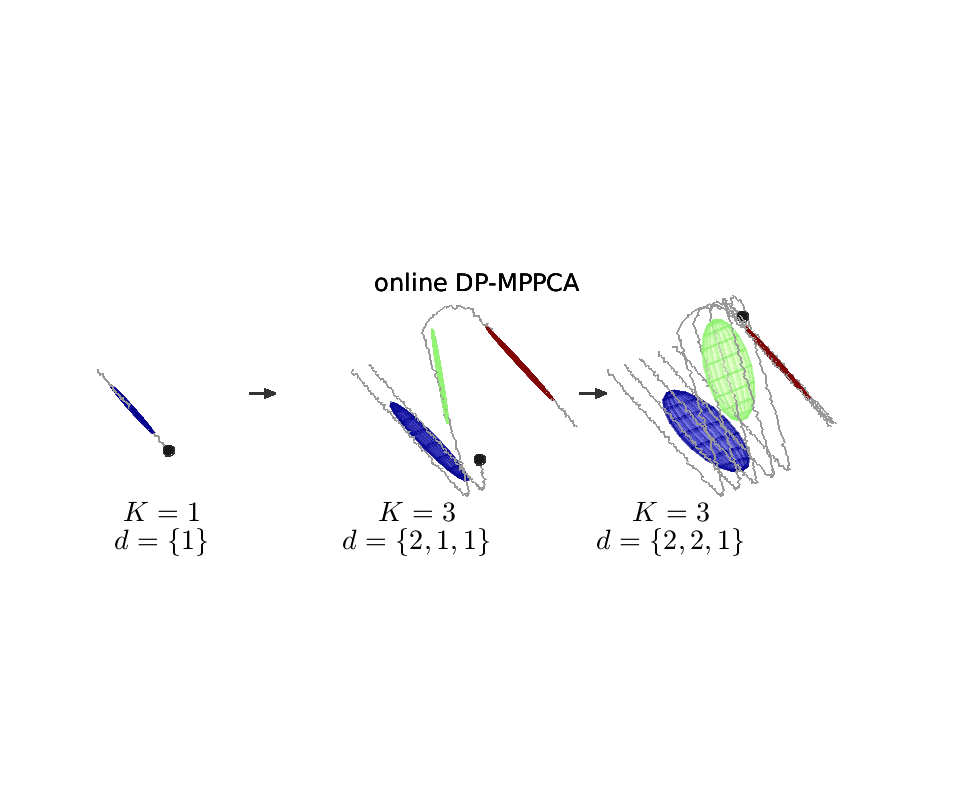}
\caption{Non-parametric online clustering of Z-shaped streaming data under small variance asymptotics with: \textit{(top)} online DP-GMM, \textit{(bottom)} online DP-MPPCA.}\label{fig: Z_Coceptual}
\end{figure}

\textbf{Loss function} $\mathcal{L}(z_{t+1},d_{t+1,z_{t+1}}, \mb{\mu}_{t+1,z_{t+1}}, \mb{U}_{t+1,z_{t+1}}^{d_{t+1,z_{t+1}}})$: The loss function optimized at time step $t+1$ is
\begin{multline*}
	\mathcal{L}(z_{t+1},d_{t+1,z_{t+1}}, \mb{\mu}_{t+1,z_{t+1}}, \mb{U}_{t+1,z_{t+1}}^{d_{t+1,z_{t+1}}}) = 
	\lambda K + \\
	\lambda_1 d_{t+1,z_{t+1}}
	+ \text{dist}(\mb{\xi}_{t+1}, \mb{\mu}_{t+1,z_{t+1}} , \mb{U}_{t+1,z_{t+1}}^{d_{t+1,z_{t+1}}})^{2} \\ 
	 \leq \mathcal{L}(z_{t+1},d_{t, z_{t+1}}, \mb{\mu}_{t, z_{t+1}}, \mb{U}_{t, z_{t+1}}^{d_{t,z_{t+1}} }).
\end{multline*} 
The loss function provides an intuitive trade-off between the fitness term $\text{dist}(\mb{\xi}_{t+1}, \mb{\mu}_{t+1,z_{t+1}}, \mb{U}_{t+1,z_{t+1}}^{d_{t+1,z_{t+1}}})^{2}$ and the model selection parameters $K$ and $d_k$. Increasing the number of clusters or the subspace dimension of the assigned cluster decreases the distance of the datapoint to the assigned subspace at the cost of penalty terms $\lambda$ and $\lambda_1$. Parameters of the assigned cluster are updated in a greedy manner such that the loss function is guaranteed to decrease at the current time step. In case a new cluster is assigned to the datapoint, the loss function at time $t$ is evaluated with the cluster having the lowest cost among the existing set of clusters. Note that setting $d_{t,i} = 0$ by choosing $\lambda_1 \gg 0$ gives the same loss function and objective function as the online DP-GMM algorithm with isotropic Gaussians.

To illustrate the difference of encoding between online DP-means and online DP-MPPCA, we evaluate the performance of the algorithms on a Z-shaped $3$-dimensional stream of datapoints with penalty parameters $\{\lambda = 35, \sigma^{2} = 100\}$ for online DP-GMM, and $\{\lambda = 14, \lambda_1 = 2, \sigma^{2} = 1, b_m = 1 \times 10^{4} \}$ for online DP-MPPCA. Fig.\ \ref{fig: Z_Coceptual} shows that online DP-GMM under small variance asymptotics fails to represent the variance in the demonstrations with $d=0$, whereas the number of clusters and the subspace dimension adequately evolves for online DP-MPPCA to model the underlying distribution.
 

%
%

\section{Online HDP-HSMM} \label{sec: online_HDP_HSMM}

In this section, we first briefly describe HSMM and its Bayesian non-parametric extension, and then present our incremental formulation to estimate the parameters of an infinite HSMM, $\Theta_{\text{HSMM}} = \Big\{\mb{\mu}_{i},\mb{\Sigma}_{i},\{a_{i,m} \}_{m=1}^{K}, \mu_{i}^{S},\Sigma_{i}^{S} \Big\}_{i=1}^{K}$,  where the output distribution of $i$-th state is represented by a parsimonious multivariate Gaussian $\mathcal{N}(\mb{\mu}_{i}, \mb{\Lambda}_{i}^{d_i} \; {\mb{\Lambda}_{i}^{d_{i}}}^\trsp+ \sigma^2\mb{I})$. Compared to the previous section, transition probabilities and an explicit state duration model for each state will be introduced as additional parameters.

\subsection{Hidden Semi-Markov Model (HSMM)}

A hidden Markov model describes a latent Markov process with transitions between a finite number of states at discrete times, and emission of an observation in each state. Spatio-temporal encoding with HMMs can handle movements with variable durations, recurring patterns, options in the movement, or partial/unaligned demonstrations. Learning in HMMs usually requires experimenting with the structure of transitions and the number of latent states. For example, left-to-right HMMs preclude all the states previously visited by setting constraints to the corresponding transition probabilities to be zero. 
HMMs implicitly assume that the duration of staying in a state follows a geometric distribution. This assumption is often limiting, especially for the modeling of sequences with long state dwell-times \citep{Rabiner89}. 

A hidden semi-Markov model (HSMM) relaxes the Markovian structure of state transitions by relying not only upon the current state but also on the duration/elapsed time in the current state. An explicit duration HSMM sets the self-transition probabilities to zero and explicitly models the state duration with a parametric distribution \citep{Yu10} (for simplicity, we use a Gaussian distribution to model the state duration, but other distributions may better model durations). Note that the HSMM extracts the spatio-temporal regularities of the demonstrations. Time is only included as a relative duration between two consecutive states in this representation. On the two sides of the spectrum, a flat duration distribution corresponds to an atemporal state, while a peak distribution corresponds to a finite-state machine with an automatic switching to the next state after a given number of time steps. The HSMM models the state duration in-between these two extremes. Moreover, situations where demonstrations are performed with large temporal variations but similar state variations (such as very fast and very slow demonstrations), we recommend using a GMM to model the state duration, as a single Gaussian would yield a high bias and high variance of the duration model in such a situation.


\subsection{Hierarchical Dirichlet Process Hidden Semi-Markov Model (HDP-HSMM)}

Specifying the number of latent states in an HMM/HSMM is often difficult. Model selection methods such as cross-validation or Bayesian Information Criterion (BIC) are typically used to determine the number of states. Bayesian non-parametric approaches comprising of HDPs provide a principled model selection procedure by Bayesian inference in an HMM/HSMM with infinite number of states. Interested readers can find details of DPs and HDPs for specifying an infinite set of conditional transition distribution priors in \cite{Teh06}.

HDP-HMM \citep{Beal02,VanGael08} is an infinite state Bayesian non-parametric generalization of the HMM with HDP prior on the transition distribution. In this model, the state transition distribution for each state follows a Dirichlet process $\mb{G}_i \sim \text{DP}(\alpha, \mb{G}_0)$ with concentration parameter $\alpha$ and shared base distribution $\mb{G}_0$, such that $\mb{G}_0$ is the global Dirichlet process $\mb{G}_0 \sim \text{DP}(\gamma,\mb{H})$ with concentration parameter $\gamma$ and base distribution $\mb{H}$. The top level DP enables sharing of the existing states with a new state created under a bottom level DP for each state and encourages visiting of the same consistent set of states in the sequence. Let $\mb{\beta}$ denote the weights of $\mb{G}_0$ in its stick-breaking construction \citep{sethuraman94}, then the non-parametric approach takes the form
\begin{align*}
	\mb{\beta}|\gamma &\;\sim\; \text{GEM}(\gamma), \\ 
	\mb{\pi}_i | \alpha, \mb{\beta} &\;\sim\; \text{DP}(\alpha, \mb{\beta}), \\  
	\{\mb{\mu}_i,\mb{\Lambda}_i^{d_i}, d_i \} &\;\sim\; \mb{H}, \\ 
	z_t &\;\sim\; \text{Mult}(\mb{\pi}_{z_{t-1}}), \\ 
	\mb{\xi}_t | z_t &\;\sim\; \mathcal{N}(\mb{\mu}_i, \mb{\Lambda}_i^{d_i} {\mb{\Lambda}_i^{d_i}}^\trsp+ \sigma^2\mb{I}),
\end{align*} 
where $\text{GEM}$ represents the Griffiths, Engen and McCloskey distribution \citep{Pitman02}. Without loss of generality, we have used here the parsimonious representation of a Gaussian for the output distribution of a state. 

\cite{Johnson13} presented an extension of HDP-HMM to HDP-HSMM by explicitly drawing the state duration distribution parameters and precluding the self-transitions. Other extensions such as sticky HDP-HMM \citep{Fox08} add a self-transition bias parameter to the DP of each state to prolong the state-dwell times. We take a simpler approach to explicitly encode the state duration by setting the self-transition probabilities to zero and estimating the parameters $\{\mu_i^{S}, \Sigma_i^{S} \}$ empirically from the hidden state sequence $\{z_1, \ldots, z_T \}$. 

Note that learning the model in this Bayesian non-parametric setting involves computing the posterior distribution over the latent state, the output state distribution and the transition distribution parameters. The problem is more challenging than the maximum likelihood parameter estimation of HMMs and requires MCMC sampling or variational inference techniques to compute the posterior distribution. Performing small-variance asymptotics of the joint likelihood of HDP-HMM, on the other hand, yields the maximum \emph{a posteriori} estimates of the parameters that iteratively minimize the loss function\footnote{Setting $d_i = 0$ by choosing $\lambda_1 \gg 0$ gives the loss function formulation with isotropic Gaussian under small variance asymptotics \citep{Roychowdhury13}.}
\begin{multline*}
	\mathcal{L} (\mb{z}, \mb{d}, \mb{\mu}, \mb{U}, \mb{a}) = 
	\sum_{t=1}^{T} \text{dist}(\mb{\xi}_t, \mb{\mu}_{z_t} , \mb{U}_{z_t}^{d_i})^{2} + \lambda(K - 1 ) \\ 
	+ \lambda_1 \sum_{i=1}^{K}d_i - \lambda_2 \sum_{t=1}^{T-1} 
	\log(a_{z_{t}, z_{t+1}}) + \lambda_3 \sum_{i=1}^{K} (\tau_{i} - 1),
\end{multline*} 
where $\lambda_2, \lambda_3 > 0$ are the additional penalty terms responsible for prolonging the state duration estimates compared to the loss function in Eq.\ \eqref{Eq: LossF-DP-MPPCA}. The $\lambda_2$ term favors the transitions to states with higher transition probability (states which have been visited more often before), $\lambda_3$ penalizes for transition to unvisited states with $\tau_{i}$ denoting the number of distinct transitions out of state $i$, and $\lambda, \lambda_1$ are the penalty terms for increasing the number of states and the subspace dimension of each output state distribution. 

\subsection{Online Inference in HDP-HSMM}

For the online setting, we denote the parameter set $\Theta_{\text{HSMM}}$ at time $t$ as $\Theta_{t, \text{HSMM}}$. Given the observation $\mb{\xi}_{t+1}$, we now present the cluster assignment and the parameter update steps for the online incremental version of HDP-HSMM.

\subsubsection{Cluster Assignment $z_{t+1}$:} 
The datapoint $\mb{\xi}_{t+1}$ is assigned to cluster $z_{t+1}$ based on the rule
\begin{align}
	z_{t+1} &= \argmin_{i=1:K+1}\begin{cases}
	q_{1,i},& \text{if } \{a_{t,z_{t}, i} > 0, i \leq K\}\\
	q_{2,i},& \text{if } \{a_{t,z_{t}, i} = 0, i \leq K\}\\
	q_{3,i}	,              & \text{otherwise,} 
	\end{cases}
	\nonumber\\
	q_{1,i} &= \text{dist}(\mb{\xi}_{t+1}, \mb{\mu}_{t,i} , \mb{U}_{t,i}^{d_i})^{2} - \lambda_2 
	\text{log}\; a_{t, z_{t}, i}, 
	\label{Eq: HDP-HSMM-q1}\\
	q_{2,i} &= \text{dist}(\mb{\xi}_{t+1}, \mb{\mu}_{t,i}, \mb{U}_{t,i}^{d_i})^{2} - \lambda_2 
	\text{log}\frac{1}{\sum_{k=1}^{K}\!c_{t, z_{t},k}\!+\!1} \!+\! \lambda_3, 
	\label{Eq: HDP-HSMM-q2}\\
	q_{3,i} &= \lambda - \lambda_2\text{log}\frac{1}{\sum_{k=1}^{K} c_{t, z_{t},k} + 1}  + \lambda_3, 
	\label{Eq: HDP-HSMM-q3}
\end{align} 
where $c_{t,i,j}$ 
is an auxiliary transition variable that counts the number of visits from state $i$ to state $j$ till time $t$. The assignment procedure evaluates the cost on two main criteria: 1) distance of the datapoint to the existing cluster subspaces given by $\text{dist}(\mb{\xi}_{t+1}, \mb{\mu}_{t,i} , \mb{U}_{t,i}^{d_i})$, and 2) transition probability of moving from the current state to the other state $a_{t, z_t,i}$. The procedure favors the next state to be one whose distance from the subspace of a cluster is low and whose transition probability is high, as seen in Eq.\ \eqref{Eq: HDP-HSMM-q1}. If the probability of transitioning to a given state is zero, an additional penalty of $\lambda_3$ is added along with a pseudo transition count to that state $\frac{1}{\sum_{k=1}^{K} c_{t,z_{t},k} + 1}$. Finally, if the cost of transitioning to a new state at subspace distance $\lambda$ in Eq.\ \eqref{Eq: HDP-HSMM-q3} is lower than the cost evaluated in Eq.\ \eqref{Eq: HDP-HSMM-q1} and Eq.\ \eqref{Eq: HDP-HSMM-q2}, a new cluster is created with the datapoint and default parameters.

\subsubsection{Parameter Updates $\Theta_{t+1,\text{HSMM}}$:} 
Given the cluster assignment $z_{t+1} = i$, we first estimate the parameters $\mb{\mu}_{t+1,i}, \mb{U}_{t+1,i}^{d_{t, i}}, d_{t+1,i},$ and $\mb{\Sigma}_{t+1,i}$ following the update rules in Eqs \eqref{Eq: Dp-means2}, \eqref{Eq: online DP-PCA}, \eqref{Eq: OnlineDPPCA-di} and \eqref{Eq: OnlineDPPCA-SigmaU}, respectively. We update the transition probabilities via the auxiliary transition count matrix with
\begin{align}
	c_{t+1, z_{t},z_{t+1}} &= c_{t, z_{t},z_{t+1}} + 1, 
	\label{Eq: HDP-HSMM-c_Count} \\ 
	a_{t+1, z_{t},z_{t+1}} &= c_{t+1,z_{t},z_{t+1}} \;/\; \sum_{k=1}^{K} c_{t+1, z_{t},k}. 
	\label{Eq: HDP-HSMM-a_Count}
\end{align} 

To update the state duration probabilities, we keep a count of the duration steps $s_t$ in which the cluster assignment is the same, i.e.,
\begin{equation}
	s_{t+1} = \begin{cases} s_t + 1, & \text{if } z_{t+1} = z_t, \\
	0, & \text{otherwise. } \end{cases}
\end{equation} 
	
Let us denote $n_{t,z_t}$ as the total number of transitions to other states from the state $z_t$ till time $t$. When the subsequent cluster assignment is different, $z_{t+1} \neq z_{t}$, the duration count is reset to zero, $s_{t+1} = 0$, the transition count to other states is incremented, $n_{t+1,z_t} = n_{t,z_t} + 1$, and the duration model parameters $\{\mu_{t+1,z_{t}}^{S}, \Sigma_{t+1, z_{t}}^{S}\}$ are updated as
\begin{align}
	\mu_{t+1,z_{t}}^{S} &= \mu_{t, z_{t}}^{S} + \frac{(s_{t} - \mu_{t, z_{t}}^{S})}{n_{t,z_{t}} + 1}, 
	\label{HDP-HSMM-muS_update}\\ 
	e_{t+1,z_{t}} &= e_{t,z_t} + (s_{t} - \mu_{t, z_{t}}^{S}) (s_{t} - \mu_{t+1, z_{t}}^{S}), \label{HDP-HSMM-e_update} \\ 
	\Sigma_{t+1, z_{t}}^{S} &= \frac{e_{t+1,z_t}}{n_{t,z_{t}}}. 
	\label{HDP-HSMM-SigmaS_update}
\end{align}  

\textbf{Loss function $\mathcal{L}(z_{t+1},d_{t+1,z_{t+1}}, \mb{\mu}_{t+1,z_{t+1}}, \mb{U}_{t+1,z_{t+1}}^{d_{t+1,z_{t+1}}}, \\ a_{t+1, z_t, z_{t+1}})$}: The parameters updated at time step $t+1$ minimize the loss function
\begin{align*}
	\mathcal{L} &
	(z_{t+1},d_{t+1,z_{t+1}}, \mb{\mu}_{t+1,z_{t+1}}, \mb{U}_{t+1,z_{t+1}}^{d_{t+1,z_{t+1}}}, a_{t+1, z_t, z_{t+1}}) =  	
	\\ 
	& \lambda (K-1) + \lambda_1 d_{t+1,z_{t+1}} - \lambda_2 \log(a_{z_t, z_{t+1}}) + \lambda_3 \;\tau_{z_{t+1}} 
	\\ & \qquad \; \qquad \; 
	+  \text{dist}(\mb{\xi}_{t+1}, \mb{\mu}_{t+1,z_{t+1}} , \mb{U}_{t+1,z_{t+1}}^{d_{t+1,z_{t+1}}})^{2} 
	\nonumber \\ &  \qquad \; \quad \; 
	\leq \mathcal{L}(z_{t+1},d_{t, z_{t+1}}, \mb{\mu}_{t, z_{t+1}}, \mb{U}_{t, z_{t+1}}^{d_{t,z_{t+1}} }, a_{t,z_t, z_{t+1}}).
\end{align*} 

A decrease of the loss function ensures that the assigned cluster parameters are updated in an optimal manner. In case a new cluster is assigned to the datapoint, the loss function at time $t$ is evaluated with the cluster having the lowest cost among the existing set of clusters.

\textbf{Remark:} Note that $\lambda_2$ encourages visiting the more influential states, and $\lambda_3$ restricts the creation of new states. We do not explicitly penalize the deviation from the state duration distribution in the cluster assignment step or the loss function, and only re-estimate the parameters of the state duration in the parameter update step. Deviation from the state duration parameters may also be explicitly penalized as shown with small variance asymptotic analysis of hidden Markov jump processes \citep{Huggins15}.

\begin{algorithm}[tbp] \caption{\small{\em {Scalable Online Sequence Clustering (SOSC)}}} \label{alg:ac}
\renewcommand{\algorithmicrequire}{\textbf{Input:}}
\begin{algorithmic}[1]

\REQUIRE $<\lambda, \lambda_1, \lambda_2, \lambda_3, \sigma^{2}, b_m >$

\textbf{procedure} \text{SOSC}
\STATE Initialize $K := 1,  \{d_{0,K}, c_{0,K,K}, \mu_{0, K}^{S}, n_{0,K}, e_K\} := 0 $
\WHILE {new $\mb{\xi}_{t+1}$ is added}
\STATE $z_{t+1} = \argmin_{i=1:K+1}\begin{cases}
q_{1,i},& \text{if } \{a_{t,z_{t}, i} > 0, i \leq K\}\\
q_{2,i},& \text{if } \{a_{t,z_{t}, i} = 0, i \leq K\}\\
q_{3,i}	,              & \text{otherwise,} 
\end{cases}$ by computing $q_{1,i}, q_{2,i}, q_{3,i}$ using Eq.\ \eqref{Eq: HDP-HSMM-q1}, \eqref{Eq: HDP-HSMM-q2}, \eqref{Eq: HDP-HSMM-q3} 
\IF {$z_{t+1} = K+1$}
\STATE $K := K+1, \quad \mb{\mu}_{t+1, K} := \mb{\xi}_t, \quad \mb{\Sigma}_{t+1,K} :=  \sigma^{2} \mb{I}$
\STATE $\{d_{t+1,K}, c_{t+1,K,K}, \mu_{old, K}^{S}, n_{t+1,K}, e_{t+1,K}\} := 0 $
\ELSE
\STATE Update $\mb{\mu}_{t+1,z_{t+1}}$ using Eq.\ \eqref{Eq: Dp-means2}
\STATE Solve $\mb{R}_{t+1, z_{t+1}}$, update $\mb{U}_{t+1,z_{t+1}}^{d_{t,z_{t+1}}}$ using Eq.\ \eqref{Eq: online DP-PCA}
\STATE Update $d_{t+1,z_{t+1}}$ using Eq.\ \eqref{Eq: OnlineDPPCA-di}
\STATE Update $\mb{\Sigma}_{t+1,z_{t+1}}$ using Eq.\ \eqref{Eq: OnlineDPPCA-SigmaU}
\ENDIF
\STATE Update $c_{t+1, z_{t},z_{t+1}}, a_{t+1, z_{t},z_{t+1}}$ using Eq.\ \eqref{Eq: HDP-HSMM-c_Count}, \eqref{Eq: HDP-HSMM-a_Count}
\IF {$z_{t+1} = z_{t}$}
\STATE $s_{t+1} := s_t + 1$
\ELSE
\STATE $s_{t+1} := 0, \qquad n_{t+1,z_{t}} := n_{t,z_{t}} + 1$
\STATE Update $\mu_{t+1,z_{t}}^{S}$ using Eq.\ \eqref{HDP-HSMM-muS_update}
\STATE Update $e_{t+1,z_t}$ using Eq.\ \eqref{HDP-HSMM-e_update}
\STATE Update $\Sigma_{t+1, z_{t}}^{S}$ using Eq.\ \eqref{HDP-HSMM-SigmaS_update} for $n_{t,z_t} > 1$
\ENDIF
\STATE $z_t := z_{t+1}$
\FOR{$i:=1$ \textbf{to} $K$}
\IF [$i \neq z_{t+1}$]{$\|\mb{\mu}_{t+1,z_{t+1}} - \mb{\mu}_{t,i} \|_2 \; < \; \lambda$ } 
\STATE Merge_Clusters$(z_{t+1},i)$
\ENDIF
\ENDFOR
\ENDWHILE 
\RETURN $\{\mb{\mu}_{t,i}, \mb{\Sigma}_{t,i}, \{a_{t,i,j} \}_{j=1}^{K}, \mu_{t,i}^{S}, \Sigma_{t,i}^{S}\}_{i=1}^{K}$
\end{algorithmic}
\end{algorithm}

\section{SOSC Algorithm} \label{sec: SOSC}

SOSC is an unsupervised non-parametric online learning algorithm for clustering time-series data. It incrementally projects the streaming data in low dimensional subspaces and maintains a history of the duration steps and the subsequent transition to other subspaces. The projection mechanism uses a non-parametric locally linear principal component analysis whose redundant dimensions are automatically discarded by small variance asymptotic analysis along those dimensions, while the spatio-temporal information is stored with an infinite state hidden semi-Markov model. During learning, if a cluster evolves such that it is closer to another cluster than the threshold $\lambda$, the two clusters are merged into one and the subspace of the dominant cluster is retained. The overall algorithm is shown in Alg.\ \ref{alg:ac} (see Extension \ref{app: MnE}-$3$ for associated codes and examples).

The algorithm yields a generative model that scales well in higher dimensions and does not require computation of numerically unstable gradients for the parameter updates at each iteration. These desirable aspects of the model comes at a cost of hard/deterministic clusters which could be a bottleneck for some applications. Non-parametric treatment aids the user to build the model online without specifying the number of clusters and the subspace dimension of each cluster, as the parameter set grows with the size/complexity of the data during learning. The penalty parameters introduced are more intuitive to specify and act as regularization terms for model selection based on the structure of the data. Note that the order of the streaming data plays an important role during learning, and multiple starts from different initial configurations may lead to different solutions as we update the model parameters after registering every new sample. Alternatively, the model parameters can be initialized with a batch algorithm after storing a few demonstrations, or the parameters can be updated sequentially in a mini-batch manner. Systematic investigation of these approaches is subject to future work.

\begin{figure*}[thp]
\centering
\includegraphics[trim={1.4cm 4.4cm 7.2cm 1.2cm},clip,scale = 0.66]{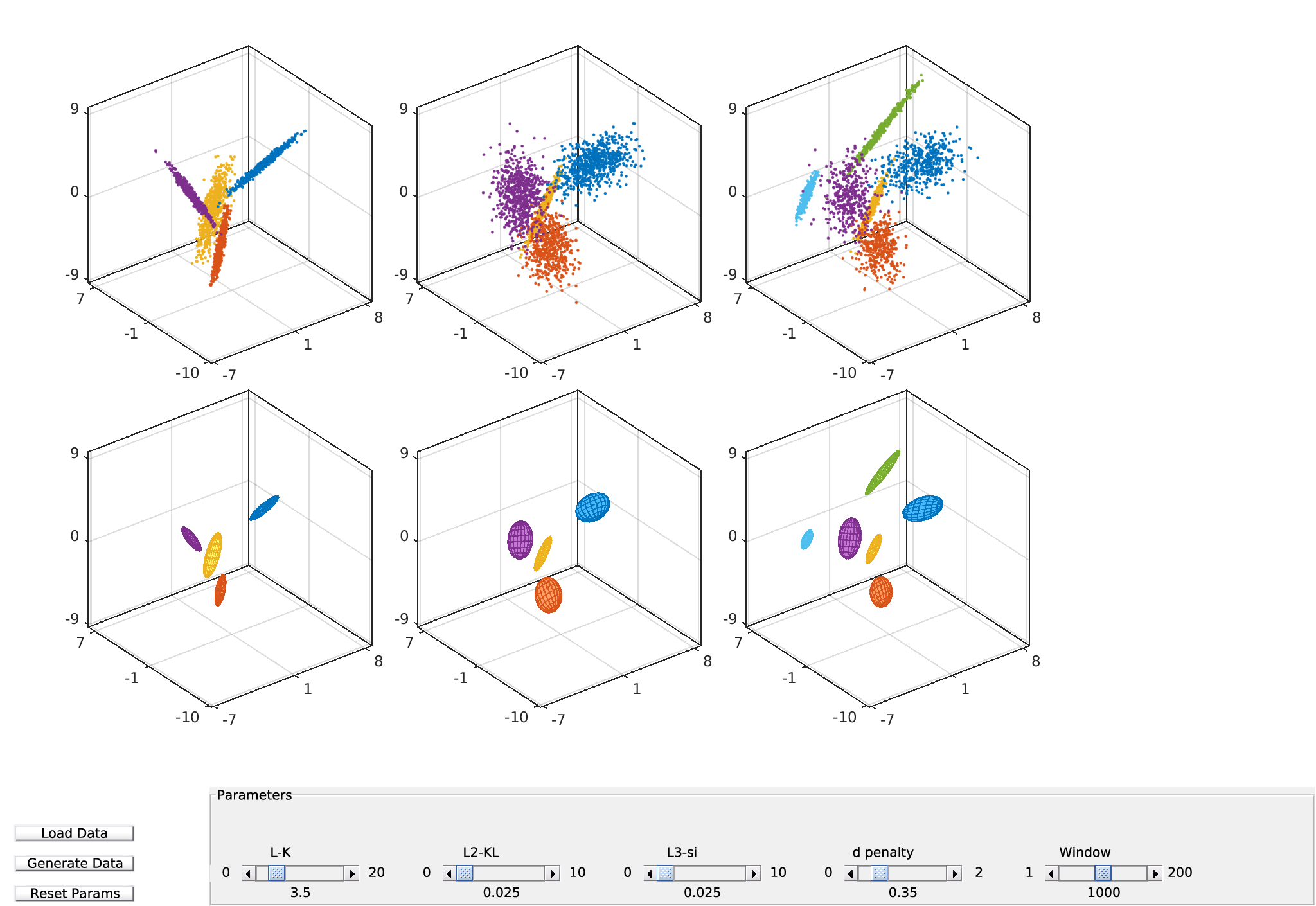} 
\caption{Non-stationary data shown on top is encoded with the SOSC model on bottom: \textit{(left)} $K:=4$, $d_k$ is randomly chosen, $t := 1 \ldots 2500$, \textit{(middle)} $K:=4$, $d_k := D - d_k$, $t:=2501 \ldots 5000$, \textit{(right)} $K:=6$, $d_k$ is the same as before, $t:=5001 \ldots 7500.$ }
\label{fig: SyntheticDataNS}
\end{figure*}

\subsection{Task-Parameterized Formulation of SOSC}

Task-parameterized models provide a probabilistic formulation to deal with different real world situations by adapting the model parameters in accordance with the external task parameters that describe the situation, instead of hard coding the solution for each new situation or handling it in an \textit{ad hoc} manner \citep{Wilson99,Calinon16,Tanwani2016a,Tanwani18}. Task-parameterized formulation of the SOSC model is able to handle new situations by defining external reference frames such as coordinate systems attached to an object whose position and orientation may change during the task. When a different situation occurs (position/orientation of the object changes), changes in the task parameters/reference frames are used to modulate the model parameters in order to adapt the robot movement to the new situation. 

We represent the task parameters with $P$ coordinate systems, defined by $\{\mb{A}_{t,j}, \mb{b}_{t,j}\}_{j=1}^{P}$, where $\mb{A}_{t,j}$ denotes the orientation of the frame as a rotation matrix and $\mb{b}_{t,j}$ represents the origin of the frame at time $t$. Each demonstration $\mb{\xi}_t$ is observed from the viewpoint of $P$ different experts/frames, with $\mb{\xi}_t^{(j)} = \mb{A}_{t,j}^{-1} (\mb{\xi}_t - \mb{b}_{t,j})$ denoting the demonstration observed with respect to frame $j$. The parameters of the task-parameterized SOSC model are defined by $\Theta_{t,\text{TP-HSMM}} = \Big\{\{\mb{\mu}_{t,i}^{(j)},\mb{\Sigma}_{t,i}^{(j)}\}_{j=1}^P, \{a_{t,i,m} \}_{m=1}^{K}, \mu_{t,i}^{S},\Sigma_{t,i}^{S} \Big\}_{i=1}^{K}$, where $\mb{\mu}_{t,i}^{(j)}$ and $\mb{\Sigma}_{t,i}^{(j)}$ define the mean and the covariance matrix of $i$-th mixture component in frame $j$ at time $t$. Parameter updates of the task-parameterized SOSC algorithm remain the same as described in Alg.\ \ref{alg:ac}, except the computation of the mean and the covariance matrix is repeated for each frame separately.

In order to fuse information from the different experts in an unseen situation represented by the frames $\{\mb{\tilde{A}}_{t,j}, \mb{\tilde{b}}_{t,j} \}_{j=1}^P$, we linearly transform the Gaussians back to the global coordinates with $\{\mb{\tilde{A}}_{t,j}, \mb{\tilde{b}}_{t,j} \}_{j=1}^P$, and retrieve the new model parameters $\{\mb{\tilde{\mu}}_{t,i}, \mb{\tilde{\Sigma}}_{t,i}\}$ for the $i$-th mixture component by computing the products of the linearly transformed Gaussians
\begin{equation}
	\label{Eq: ProdGaussEq} 
	\mathcal{N}(\mb{\tilde{\mu}}_{t,i}, \mb{\tilde{\Sigma}}_{t,i}) \;\propto\;
  \prod\limits_{j=1}^P \mathcal{N}\left(\mb{\tilde{A}}_{t,j} \mb{\mu}_{t,i}^{(j)} + 
  \mb{\tilde{b}}_{t,j}, \mb{\tilde{A}}_{t,j} \mb{\Sigma}_{t,i}^{(j)} \mb{\tilde{A}}_{t,j}^{\trsp} \right). 
\end{equation} 

The product of Gaussians can be evaluated in an analytical form with
\begin{align}
	\mb{\tilde{\mu}}_{t,i} &= \mb{\tilde{\Sigma}}_{t,i} \sum_{j=1}^{P}\left(\mb{\tilde{A}}_{t,j} \mb{\Sigma}_{t,i}^{(j)} 
	\mb{\tilde{A}}_{t,j}^{\trsp}\right)^{-1} \left(\mb{\tilde{A}}_{t,j} \mb{\mu}_{t,i}^{(j)} +  \mb{\tilde{b}}_{t,j} \right), 
	\nonumber\\ 
	\mb{\tilde{\Sigma}}_{t,i} &=  \left( \sum_{j=1}^{P} 
	\left(\mb{\tilde{A}}_{t,j} \mb{\Sigma}_{t,i}^{(j)} \mb{\tilde{A}}_{t,j}^{\trsp}\right)^{-1} \right)^{-1}. 
	\label{Eq: GeneralizedSigma}
\end{align} 

\textbf{Loss function $\mathcal{L} (z_{t+1}, \tilde{d}_{t+1,z_{t+1}}, \mb{\tilde{\mu}}_{t+1,z_{t+1}}, \mb{\tilde{U}}_{t+1,z_{t+1}}^{\tilde{d}_{t+1,z_{t+1}}}, \\ a_{t+1, z_t, z_{t+1}})$:}
Under the small variance asymptotics, the loss function at time step $t+1$ for the task-parametrized SOSC model with the resulting $\mathcal{N}(\mb{\tilde{\mu}}_{t+1,z_{t+1}}, \mb{\tilde{\Sigma}}_{t+1,z_{t+1}})$ yields
\begin{multline*}
	\mathcal{L} (z_{t+1}, \tilde{d}_{t+1,z_{t+1}}, \mb{\tilde{\mu}}_{t+1,z_{t+1}}, 
	\mb{\tilde{U}}_{t+1,z_{t+1}}^{\tilde{d}_{t+1,z_{t+1}}}, a_{t+1, z_t, z_{t+1}}) =  
	\\  
	\lambda (K-1) + \lambda_1 \tilde{d}_{t+1,z_{t+1}} - \lambda_2 \log(a_{z_t, z_{t+1}}) + \lambda_3 \;\tau_{z_{t+1}} 
	\\ 
	+  \text{dist}(\mb{\xi}_{t+1}, \mb{\tilde{\mu}}_{t+1,z_{t+1}} , \mb{\tilde{U}}_{t+1,z_{t+1}}^{\tilde{d}_{t+1,z_{t+1}}})^{2} 	
	\\  
	\leq 
	\mathcal{L}(z_{t+1},\tilde{d}_{t, z_{t+1}}, \mb{\tilde{\mu}}_{t, z_{t+1}}, 
	\mb{\tilde{U}}_{t, z_{t+1}}^{\tilde{d}_{t,z_{t+1}} }, a_{t,z_t, z_{t+1}}),
\end{multline*} 
where $\mb{\tilde{U}}_{t+1,z_{t+1}}^{\tilde{d}_{t+1,z_{t+1}}}$ corresponds to the basis vectors of the resulting $\mb{\tilde{\Sigma}}_{t+1,z_{t+1}}$ and $\tilde{d}_{t+1,z_{t+1}} = \min_{j} d_{t+1,z_{t+1}}^{(j)}$, i.e., the product of Gaussians subspace dimension is defined by the minimum of corresponding subspace dimensions of the Gaussians in $P$ reference frames for the $z_{t+1}$ mixture component.

\section{Experiments, Results and Discussion}\label{sec: Experiments}

In this section, we first evaluate the performance of the SOSC model to encode the synthetic data with a $3$-dimensional illustrative example, followed by its capability to scale in high dimensional spaces. We then consider a real-world application of learning robot manipulation tasks for semi-autonomous teleoperation with the proposed task-parameterized SOSC algorithm. The goal is to assess the performance of the SOSC model to handle noisy online time-series data in a parsimonious manner. 

\subsection{Synthetic Data}

\subsubsection{Non-Stationary Learning with $3$-Dimensional Data:} 
We consider a $3$-dimensional stream of datapoints $\mb{\xi}_t \in \mathbb{R}^{3}$ generated by stochastic sampling from a mixture of clusters that are connected in a left-right cyclic HSMM. The centers of the clusters are successively drawn from the interval $[-5,5]$ such that the next cluster is at least $4 \sqrt{D}$ units farther than the existing set of clusters. Subspace dimension of each cluster is randomly chosen to lie up to $(D-1)$ dimensions (a line or a plane for $3$-dimensions), and the basis vectors are sampled randomly in that subspace. Duration steps in a given state are sampled from a uniform distribution in the interval $[70,90]$ after which the data is subsequently generated from the next cluster in the model in a cyclic manner. A white noise of $\mathcal{N}(\mb{0},0.04\mb{I})$ is added to each sampled datapoint. Model learning is divided in three stages: 1) for the first $2500$ instances, the number of clusters is set to $4$ and the subspace dimension of each cluster is fixed, 2) for the subsequent $2500$ instances, we change the subspace dimension of each cluster to $(D-d_k)$ for $k = 1 \ldots K$ (for example, a line becomes a plane), while keeping the same number of clusters, and 3) two more clusters are then added in the mixture model for the next $2500$ instances without any change in the subspace dimension of the previous clusters. The parameters are defined as $\{\lambda = 3.6, \lambda_1 = 0.35, \lambda_2 = \lambda_3 = 0.025, \sigma^{2} = 0.15, b_m = 50\}$. The weights of the parameter update are based on eligibility traces as in Eq.\ \eqref{Eq: DP-PCA-ELTR} with a discount factor of $0.995$.

\begin{figure}[tbp]
\centering
\includegraphics[trim={1.3cm 1.5cm 3.8cm 3.5cm},clip,scale = 0.47]{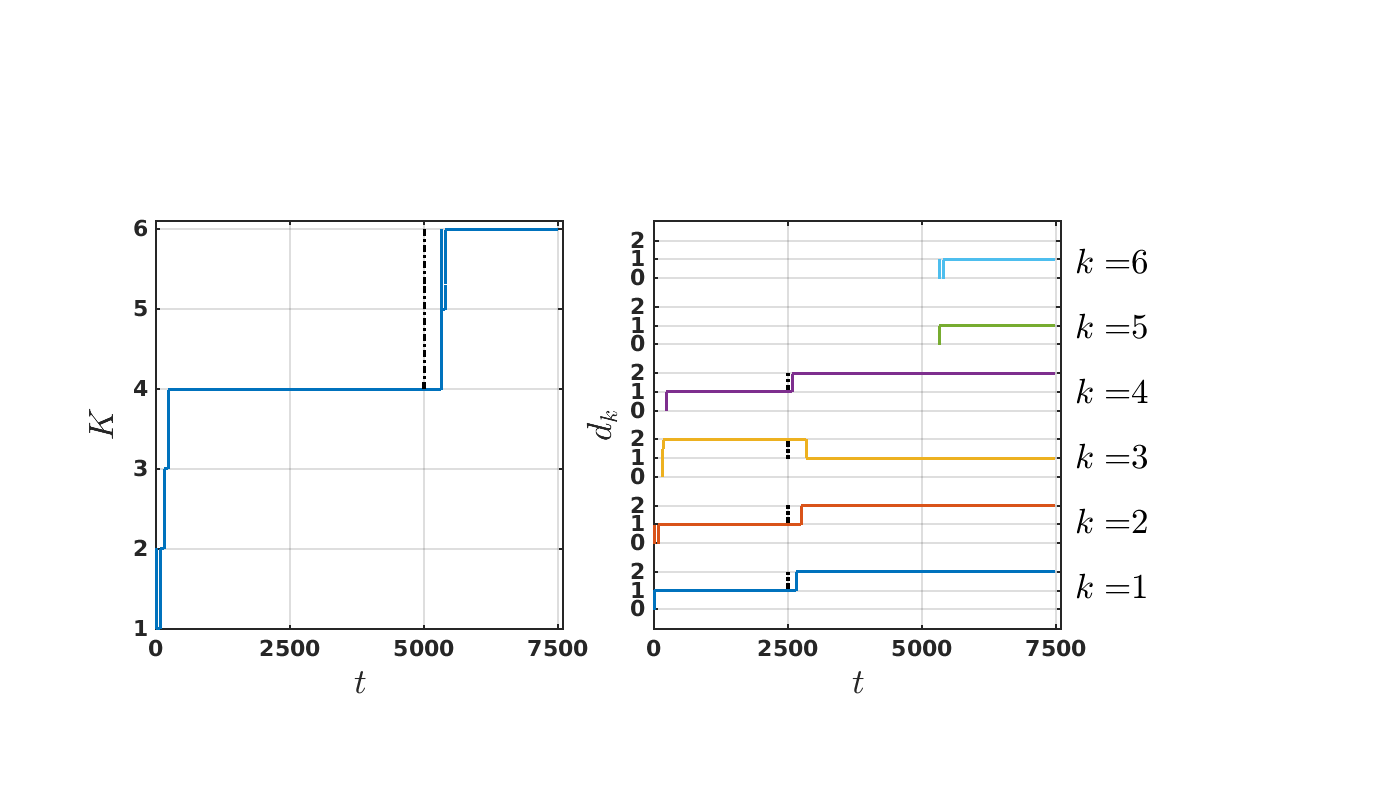} 
\caption{Evolution of $K$ and $d_k$ with number of instances.}
\label{fig: SyntheticDataNSEvo}
\end{figure}

\begin{figure}[tbp]
\centering
\includegraphics[trim={2.0cm 0.78cm 1.4cm 1.5cm},clip,scale = 0.55]{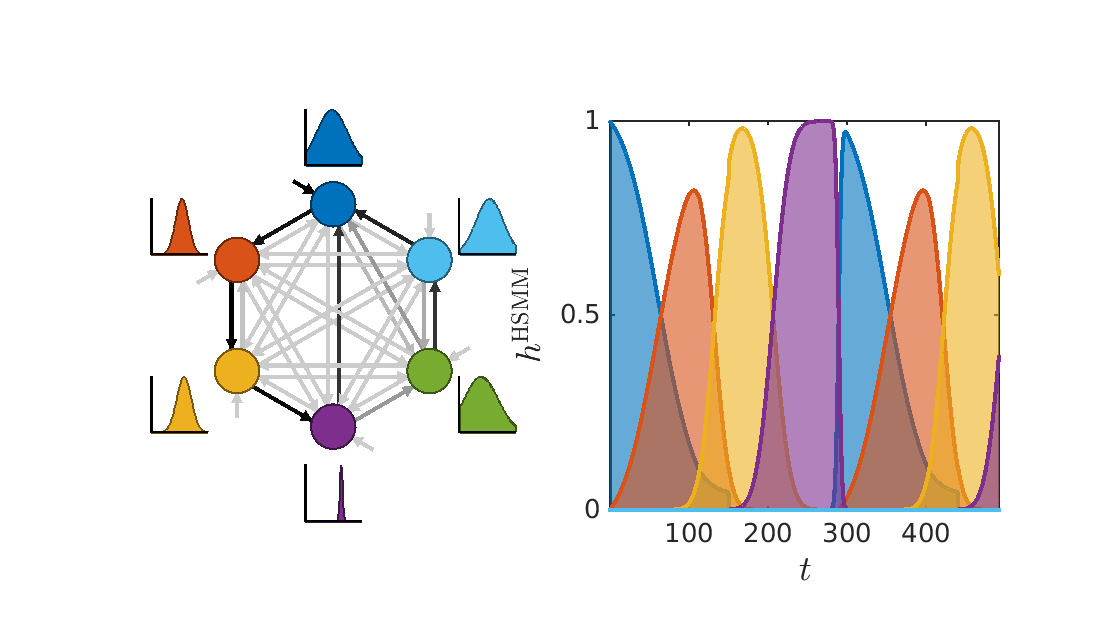} 
\caption{\textit{(left)} Learned HSMM transition matrix and state duration model representation with $s^{\max} = 150$, \textit{(right)} rescaled forward variable, $h^{\tp{HSMM}}_{t,i} = \frac{\alpha^{\tp{HSMM}}_{t,i}}{\sum_{k=1}^K \alpha^{\tp{HSMM}}_{t,k}}$, sampled from initial position.}
\label{fig: SyntheticDataNSHSMM}
\end{figure}

Results of the learned model are shown in Fig.\ \ref{fig: SyntheticDataNS}. We can see that the SOSC model is able to efficiently encode the number of clusters and the subspace dimension of each cluster in each stage of the learning process. The model projects each datapoint in the subspace of the nearest cluster contrary to the $K$-means clustering which assigns the datapoint to the nearest cluster based on the Euclidean distance metric only. The model is able to adapt the subspace dimension of each cluster in the second stage of the learning process and subsequently incorporate more clusters in the final stage with the non-stationary data. Fig.\ \ref{fig: SyntheticDataNSEvo} shows the evolution of the number of clusters and the subspace dimension of each cluster with the streaming data. Note that the encoding problem is considerably hard here as the model starts with one cluster only and adapts during the learning process. Clusters that evolve to come closer to a certain threshold are merged during the learning process. Fig.\ \ref{fig: SyntheticDataNSHSMM} shows the graphical model representation of the learned HSMM with the state transitions and the state duration model, along with a sample of the forward variable generated from the initial position (see Eq.\ \eqref{Eq. HSMM_ForwardVariable}).

\subsubsection{Stationary Learning with High-Dimensional Data:} 
\begin{figure*}[btp]
\centering
\includegraphics[trim={1.8cm 0.6cm 3cm 0.7cm},clip,scale = 0.59]{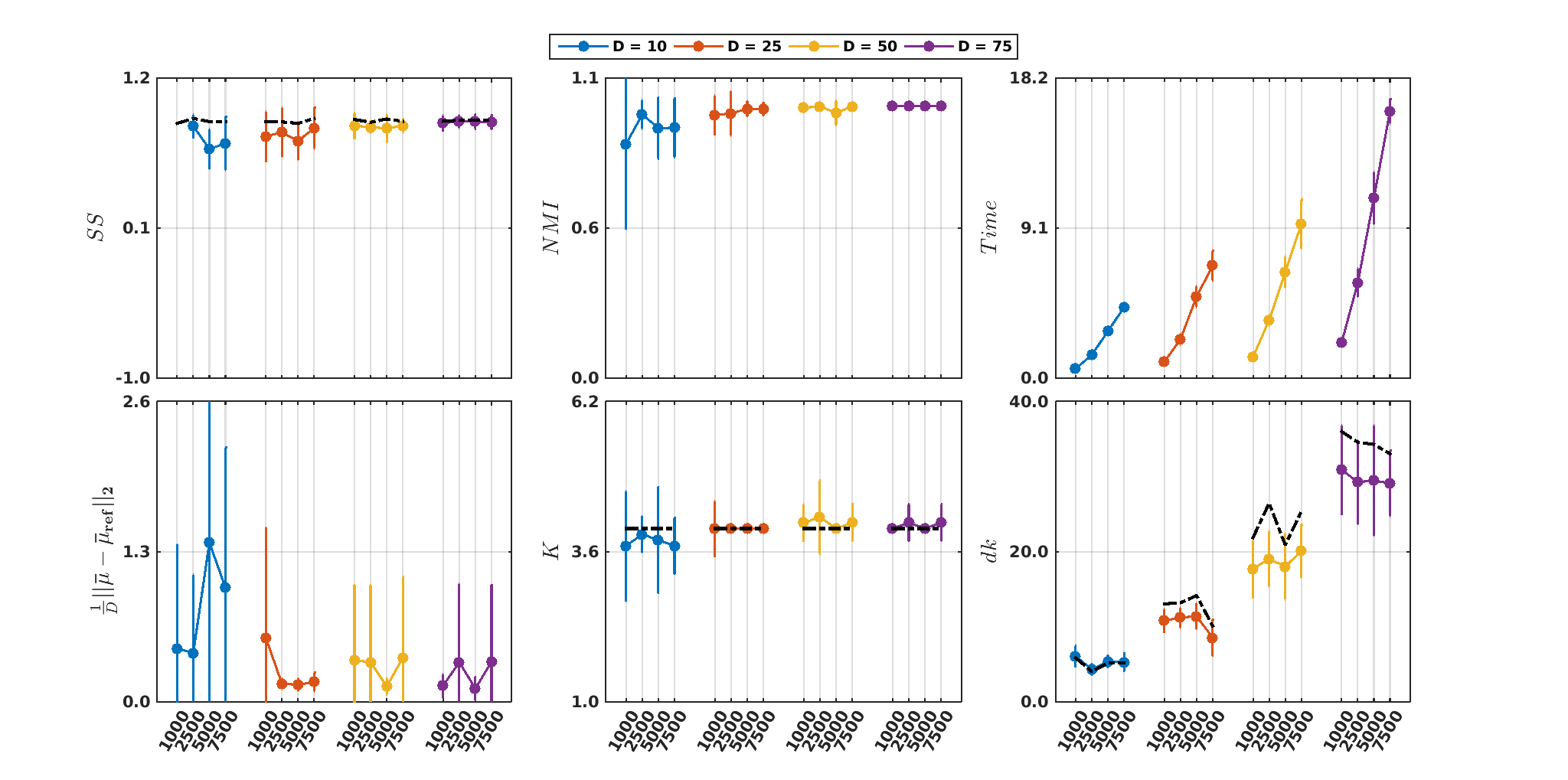} 
\caption{SOSC model evaluation to encode synthetic high-dimensional data. Results are averaged over $10$ iterations. Black dotted lines indicate the reference value: \textit{(top-left)} silhouette score (SS), \textit{(top-middle)} normalized mutual information score (NMI), \textit{(top-right)} time in seconds, \textit{(bottom-left)} average distance between learned cluster means and ground truth, \textit{(bottom-middle)} number of clusters, \textit{(bottom-right)} average subspace dimension across all clusters.}
\label{fig: SyntheticDataHD}
\end{figure*}

In this experiment, we sample the data from a stationary distribution corresponding to the first stage of the previous example where $K=4$ and the subspace of each cluster does not change in the streaming data. Dimensionality of the data is successively chosen from the set $D = \{10, 25, 50, 75\}$, and the number of instances are varied for each dimension from the set $T = \{1000, 2500, 5000, 7500 \}$. Parameter $\lambda$ is experimentally selected for each dimension to achieve satisfying results and the weights of the parameter update are linearly incremented for each cluster. Fig.\ \ref{fig: SyntheticDataHD} shows the performance of the SOSC model to encode data in high dimensions averaged over $10$ iterations. Our results show that the algorithm yields a compact encoding, as indicated by high values of the average \textit{silhouette score} (SS),\footnote{Silhouette score (SS) measures the tightness of a cluster relative to the other clusters without using any labels, $$\text{SS}_i \triangleq \frac{b_i - a_i}{\max\{a_i,b_i\}}, \qquad \text{SS}_i \in [-1,1], $$ where $a_i$ is the mean distance of $\mb{\xi}_i$ to the other points in its own cluster, and $b_i$ is the mean distance of $\mb{\xi}_i$ to the points in the closest `neighbouring' cluster.} and the \textit{normalized mutual information} (NMI) score,\footnote{Normalized mutual information (NMI) is an extrinsic information-theoretic measure to evaluate the alignment between the assigned cluster labels $\mathcal{Z}$ and the ground truth cluster labels $\mathcal{X}$, $$\text{NMI}(\mathcal{Z}, \mathcal{X}) \triangleq \frac{I(\mathcal{Z}, \mathcal{X})}{[H(\mathcal{Z}) + H(\mathcal{X}))]/2}, \quad \text{NMI}(\mathcal{Z}, \mathcal{X}) \in [0,1],$$ where $I(\mathcal{Z}, \mathcal{X})$ is the mutual information and $H(\mathcal{X})$ is the entropy of cluster labels $\mathcal{X}$.} while being robust to the intrinsic subspace dimension of the data and the number of clusters.



\begin{figure}[tbp]
\begin{center}
\includegraphics[trim={4.0cm 1.5cm 2.0cm 1.0cm},clip,width=75mm,height=44mm]{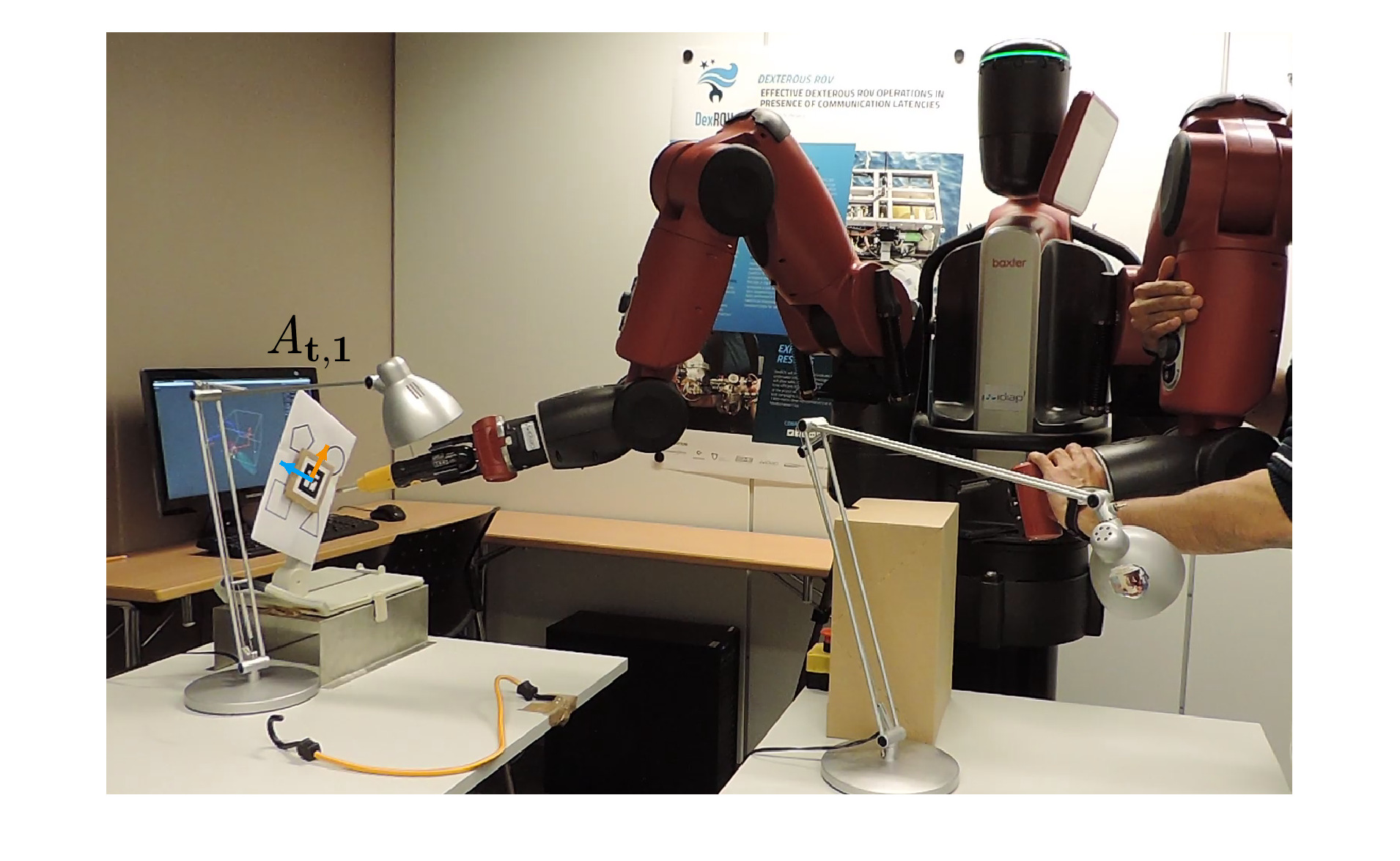} \\
\includegraphics[trim={2.3cm 1.9cm 2.0cm 1.0cm},clip,width=75mm,height=46mm]{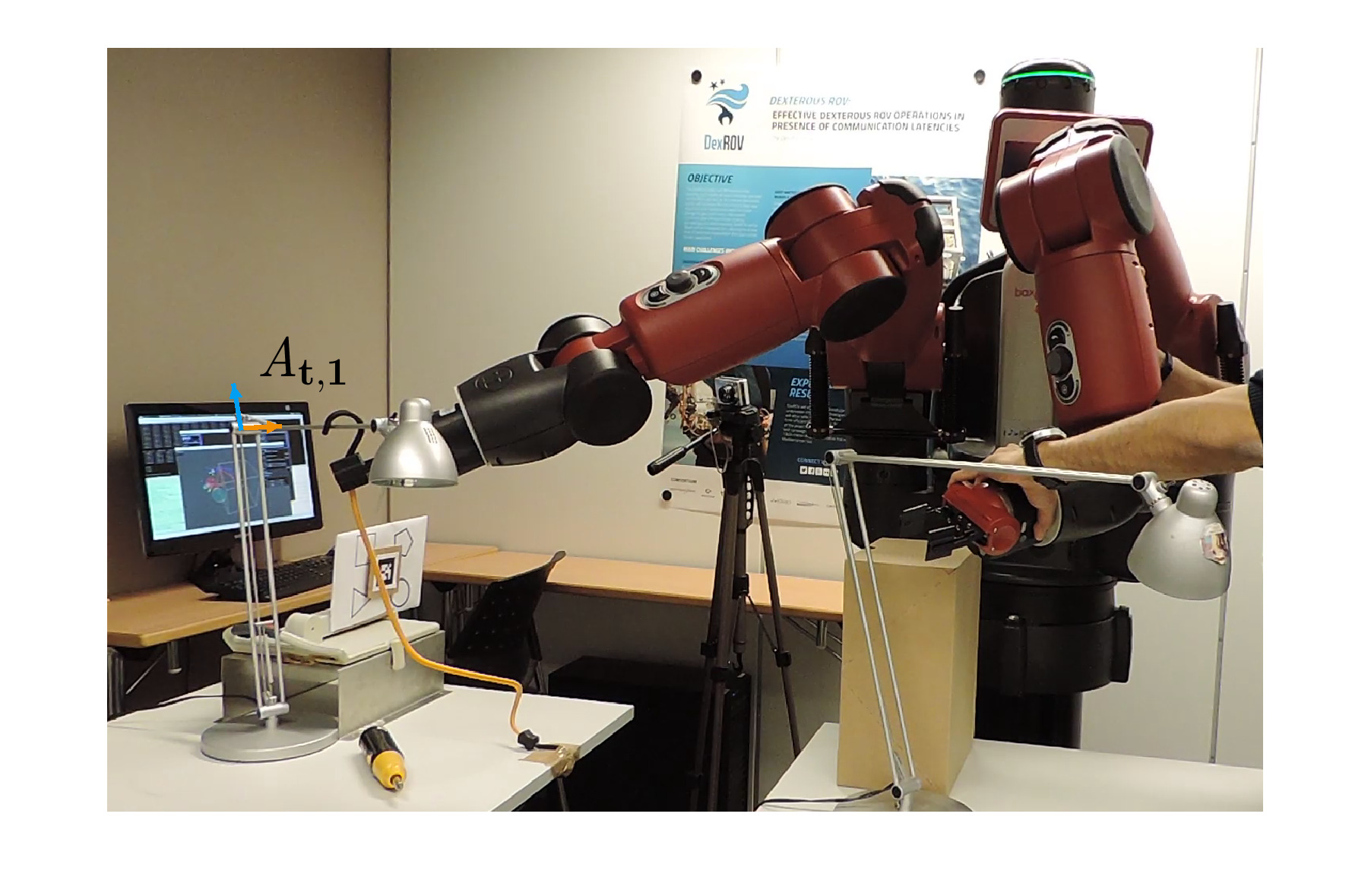} 
\end{center}
\caption{Semi-autonomous teleoperation with the Baxter robot for guided assistance of manipulation tools: \textit{(top)} screwdriving with a reference frame attached to the movable target, \textit{(bottom)} hooking a carabiner with a reference frame attached to a rotatable rod.}\label{fig: TeleopSetup}
\end{figure}

\begin{figure}[tbp]
\begin{center}
\includegraphics[trim={1.8cm 3.7cm 1.8cm 3.7cm},clip,scale = 0.55]{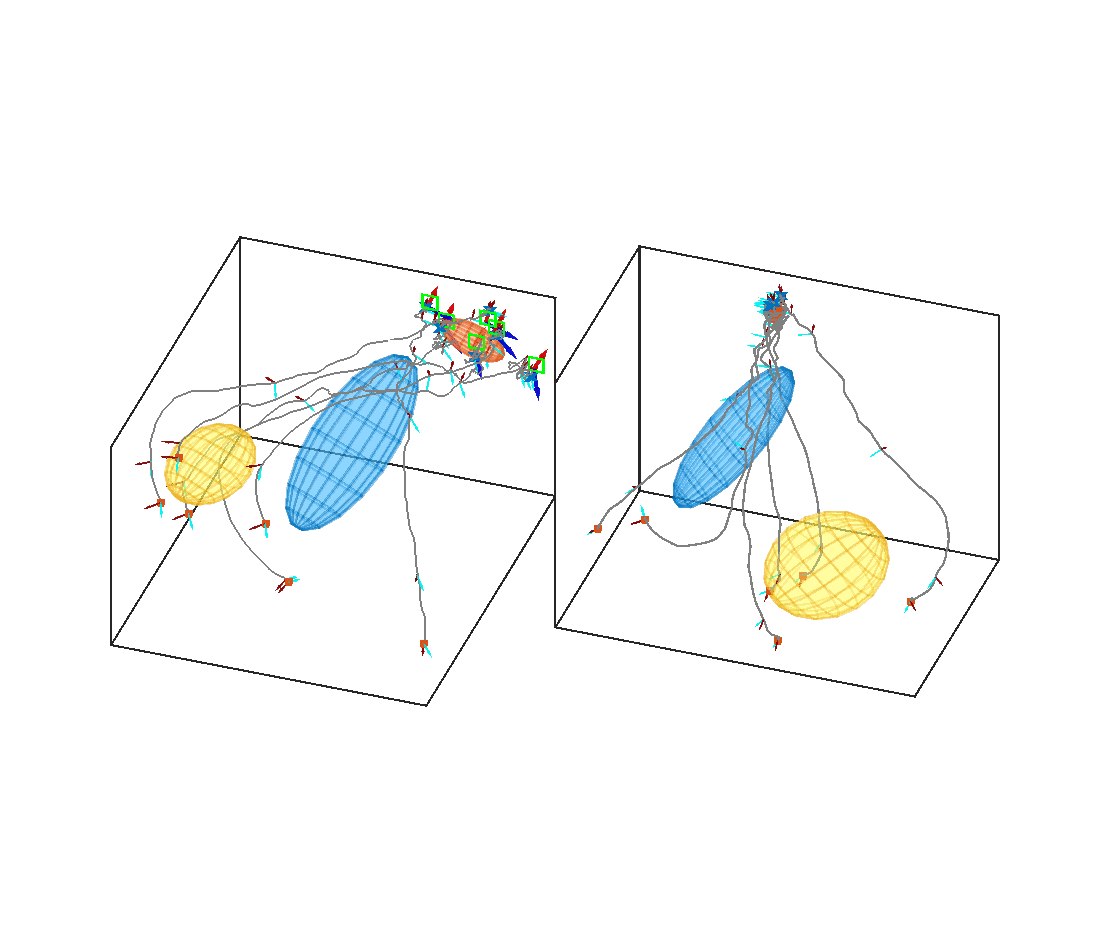} \\
\includegraphics[trim={1.8cm 3.7cm 1.8cm 3.7cm},clip,scale = 0.55]{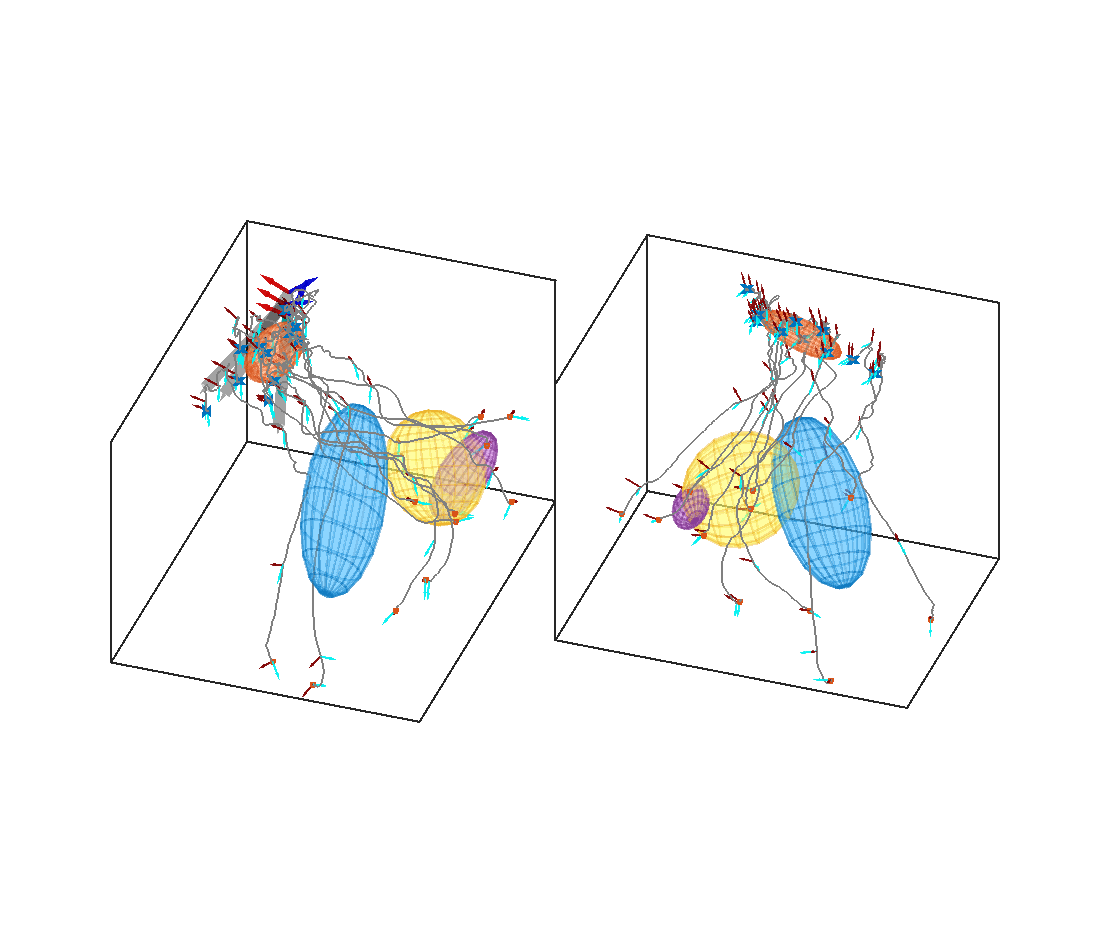}
\end{center}
\caption{Joint distribution of the task-parameterized SOSC model for guided assistance in the screwdriving task \textit{(top)} and hooking a carabiner task \textit{(bottom)}. Demonstrations and model with respect to the input dimensions of the reference frame on \textit{(left)}, and with respect to the output dimensions of the reference frame on \textit{(right)}.}
\label{fig: JointDistributionSOSC}
\end{figure}



\subsection{Learning Manipulation Skills for Semi-Autonomous Teleoperation} 

We are interested in performing remote manipulation tasks with robots via teleoperation within the DexROV project \citep{Gancet15, Gancet16}. Direct teleoperation, where the teleoperator actions are directly reproduced on the remote robot, is often infeasible due to the presence of communication latencies and noise in the feedback. Predicting/correcting the response of the operator can assist the teleoperator in executing these manipulation tasks \citep{Dragan13, Maeda15}. In this paper, we build the task-parameterized SOSC model online from the teleoperator demonstrations and provide a probabilistic formulation to predict his/her intention while performing the task. The model is used to recognize the intention of the teleoperator, and synthesize motion on the remote end to perform manipulation tasks in a semi-autonomous manner. Two didactic examples of manipulation tasks are incrementally learned for guided assistance: target tracking with a screwdriver and hooking a carabiner, see also \citep{Havoutis16} for an application of this work to hot-stabbing task.

\subsubsection{Experimental Setup:}
In our experimental setup with the Baxter robot, the operator teleoperates the right arm, with the left arm used as input device. The tool (screwdriver/carabiner) is mounted on the end-effector of the right arm and the target (movable object/rotatable rod) is placed at a reachable location from the arm. Demonstrations are performed in the direct control mode where the desired pose of the right arm is computed by adding an offset in the lateral direction to the end-effector position of the teleoperator's arm. The teleoperator guides the tool to different target locations by visual feedback, as shown in Fig.\ \ref{fig: TeleopSetup}.

\subsubsection{Learning Problem:} \label{SecLP}
Let us denote $\mb{\xi}_t = \begin{bmatrix} \mb{\xi}_t^{\ty{I}^{\trsp}} \; \mb{\xi}_t^{\ty{O}^{\trsp}} \end{bmatrix}^{\trsp}$ with $\mb{\xi}_t^{\ty{I}} \in \mathbb{R}^{7}$ and $\mb{\xi}_t^{\ty{O}} \in \mathbb{R}^{7}$ representing respectively the input state of the teleoperator arm and the input state of the teleoperator arm observed in the reference frame of the target pose of the tool (screwdriver/carabiner). The state of the teleoperator arm is represented by the position ${\mb{x}_t^{p}} \in \mathbb{R}^{3}$ and the orientation ${\mb{\varepsilon}_t^{o}} \in \mathbb{R}^{4}$ of the teleoperator arm end-effector in their respective reference frames with $D = 14$. We attach a frame $\{\mb{A}_{t,1}, \mb{b}_{t,1}\}$ to the target pose of the tool, described by
\begin{equation}
	\mb{A}_{t,1} = \begin{bmatrix}
	\mb{I}^{\ty{I}} 		& 	\mb{0}  & \mb{0}\\ 
	\mb{0}	&  	 \mb{R}_{t,1}^{\ty{O}} & \mb{0} \\
	\mb{0}	&  	 \mb{0} & \mathbf{\mathcal{E}}_{t,1}^{\ty{O}} 
	 		\end{bmatrix},
		\mb{b}_{t,1} = \begin{bmatrix}
	\mb{0} \\
	\mb{p}_{t,1}^{\ty{O}}\\
	\mb{0} 
	\end{bmatrix},
\end{equation} 
where $\mb{p}_{t,1}^{\ty{O}} \in \mathbb{R}^3, \mb{R}_{t,1}^{\ty{O}}\in \mathbb{R}^{3 \times 3}, \mathbf{\mathcal{E}}_{t,1}^{\ty{O}} \in \mathbb{R}^{4 \times 4}$ denote the Cartesian position, the rotation matrix and the quaternion matrix of the frame/tool at time $t$ respectively. Note that the frame has two components, the input component represents the teleoperator pose in the global reference frame corresponding to $\mb{\xi}_t^{\ty{I}}$, while the output component maps the teleoperator state with respect to the target pose corresponding to $\mb{\xi}_t^{\ty{O}}$. The observation variable $\mb{\xi}_t$ is augmented to couple the movement of the robot arm and the target, i.e., we learn the mapping between the teleoperator pose and the teleoperator pose observed in the reference frame of the target as a joint distribution. The coupling strength depends upon the variations observed in the demonstrations. Parts of the movement with invariant characteristics appear when approaching the target where the teleoperator movement is synchronized with the target. Based on the learned joint distribution of the task-parameterized SOSC model, we seek to recognize the intention of the teleoperator and subsequently correct the current state of the teleoperated arm. In case of communication disruptions, we solicit the model to generate movement on the remote arm in an autonomous manner until further communication is re-established. We present two formulations of the algorithm to assist the teleoperator in performing remote manipulation tasks \cite{Tanwani17}: 1) \textit{time-independent shared control}, and 2) \textit{time-dependent autonomous control}.


\textbf{Time-Independent Shared Control: } We seek to leverage upon the SOSC model to adjust the movement of the robot in following the teleoperator state in a time-independent manner based on the principle of shared control. Given the current state of the teleoperator arm $\mb{\xi}_t^{\ty{I}}$ and the task-parametrized SOSC model encoding the joint distribution as $\mathcal{N}(\mb{\tilde{\mu}}_{t,i}, \mb{\tilde{\Sigma}}_{t,i})$, the conditional probability distribution of the teleoperator arm with respect to the target $\mathcal{P}(\mb{\xi}^{\ty{O}}_t|\mb{\xi}^{\ty{I}}_t)$ can be approximated as $\mathcal{N}(\mb{\tilde{\mu}}_t^{\ty{O}}, \mb{\tilde{\Sigma}}_t^{\ty{O}})$ using Gaussian mixture regression \citep{Ghahramani94}, namely
\begin{gather}
	\mb{\tilde{\mu}}_t^{\ty{O}} = \sum_{i=1}^K h_i(\mb{\xi}^{\ty{I}}_t) \; \hat{\mb{\mu}}^{\ty{O}}_{t,i}(\mb{\xi}^{\ty{I}}_t), 
	\\ 
	\mb{\tilde{\Sigma}}_{t}^{\ty{O}} = \sum_{i=1}^{K} h_i(\mb{\xi}^{\ty{I}}_t) \Big( \hat{\mb{\Sigma}}_{t,i}^{\ty{O}} + 
	\hat{\mb{u}}_{t,i}^{\ty{O}}(\mb{\xi}^{\ty{I}}_t) {\big(\hat{\mb{\mu}}_{t,i}^{\ty{O}}(\mb{\xi}^{\ty{I}}_t)\big)}^\trsp \Big) 
	- \mb{\tilde{\mu}}^{\ty{O}}_t {\mb{\tilde{\mu}}^{\ty{O}}_t}^\trsp, \\
	\mathrm{with} \quad h_i(\mb{\xi}^{\ty{I}}_t) 
	=
	\frac{\pi_i \mathcal{N}(\mb{\xi}^{\ty{I}}_t |\; \mb{\tilde{\mu}}^{\ty{I}}_{t,i},\mb{\tilde{\Sigma}}^{\ty{I}}_{t,i})}
	{\sum_k^K \pi_k \mathcal{N}(\mb{\xi}^{\ty{I}}_t |\; \mb{\tilde{\mu}}^{\ty{I}}_{t,k},\mb{\tilde{\Sigma}}^{\ty{I}}_{t,k})}, \\
	\hat{\mb{\mu}}^{\ty{O}}_{t,i}(\mb{\xi}^{\ty{I}}_t) =
	\tilde{\mb{\mu}}^{\ty{O}}_{t,i} + \mb{\tilde{\Sigma}}^{\ty{OI}}_{t,i}{\mb{\tilde{\Sigma}}^{\ty{I}}_{t,i}}^{-1}
	(\mb{\xi}^{\ty{I}}_t-\mb{\tilde{\mu}}^{\ty{I}}_{t,i}) ,
	\label{eq:AbTP}\\
	\hat{\mb{\Sigma}}^{\ty{O}}_{t,i} 
	= \mb{\tilde{\Sigma}}^{\ty{O}}_{t,i} - 
	\mb{\tilde{\Sigma}}^{\ty{OI}}_{t,i}{\mb{\tilde{\Sigma}}^{\ty{I}}_{t,i}}^{-1} \mb{\tilde{\Sigma}}^{\ty{IO}}_{t,i},  
\end{gather} 
where $\mb{\tilde{\mu}}_{t,i}=\begin{bmatrix}\mb{\tilde{\mu}}^{\ty{I}}_{t,i} \\  \mb{\tilde{\mu}}^{\ty{O}}_{t,i}\end{bmatrix}$, and $  \mb{\tilde{\Sigma}}_{t,i}=\begin{bmatrix}\mb{\tilde{\Sigma}}^{\ty{I}}_{t,i} \quad \mb{\tilde{\Sigma}}^{\ty{I O}}_{t,i} \\ \mb{\tilde{\Sigma}}^{\ty{O I}}_{t,i} \quad \mb{\tilde{\Sigma}}^{\ty{O}}_{t,i}\end{bmatrix}$.

The output Gaussian $\mathcal{N}(\mb{\tilde{\mu}}_t^{\ty{O}}, \mb{\tilde{\Sigma}}_t^{\ty{O}})$ predicts the teleoperator state and the uncertainty associated with the state in the reference frame of the target. Let us denote $\kappa^{2}\mb{I}$ as the uncertainty associated with the teleoperator input state $\mb{\xi}_t^{\ty{I}}$ in performing the manipulation task. Higher values of $\kappa^{2}$ are used when the confidence of the teleoperator in performing the task is low, for example, when the feedback is noisy, the task is complex or the teleoperator is novice, and vice versa for lower values of $\kappa^2$. Hence, the Gaussian $\mathcal{N}(\mb{\xi}_t^{\ty{I}},\kappa^{2}\mb{I})$ represents the uncertainty associated with the current teleoperator state, while the Gaussian $\mathcal{N}(\mb{\tilde{\mu}}_t^{\ty{O}}, \mb{\tilde{\Sigma}}_t^{\ty{O}})$ predicts the teleoperator state based on the invariant patterns observed in the demonstrations with respect to the target. The resulting desired state $\mathcal{N}(\mb{\hat{\mu}}_{t}, \mb{\hat{\Sigma}}_{t})$ is obtained by taking the product of Gaussians corresponding to the teleoperator state and the predicted state, namely
\begin{equation}
	\mathcal{N}(\mb{\hat{\mu}}_{t}, \mb{\hat{\Sigma}}_{t}) \;\propto\; \mathcal{N}(\mb{\xi}_t^{\ty{I}},\kappa^{2}\mb{I}) \; 
	\mathcal{N}(\mb{\tilde{\mu}}_t^{\ty{O}}, \mb{\tilde{\Sigma}}_t^{\ty{O}}).
\end{equation} 
The resulting desired state is followed in a smooth manner with an infinite horizon linear quadratic regulator \citep{borrelli11} (see Appx.\ \ref{app: LQR_IH}).

\begin{figure}[tbp]
\begin{center}
\includegraphics[trim={1.8cm 4.25cm 1.8cm 4.2cm},clip,scale = 0.55]{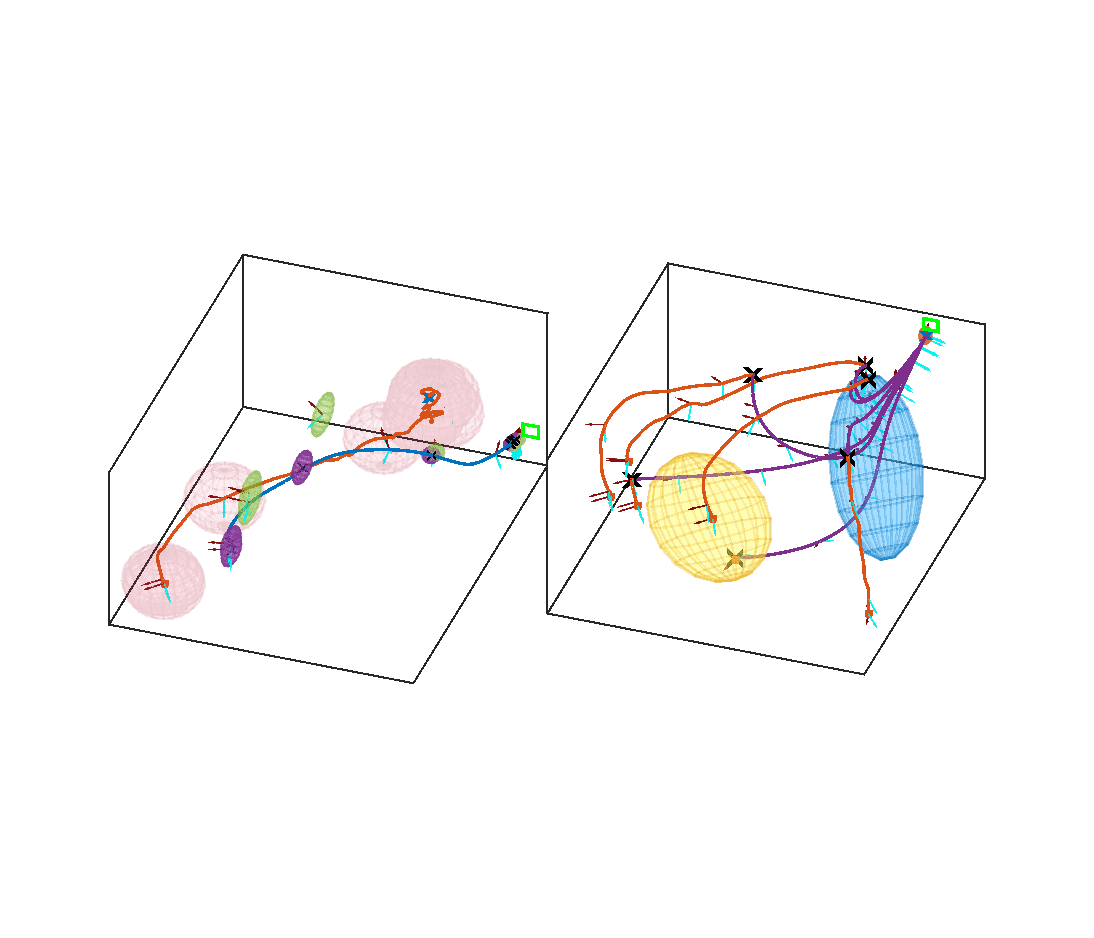} \\
\includegraphics[trim={1.8cm 4.25cm 1.8cm 4.2cm},clip,scale = 0.55]{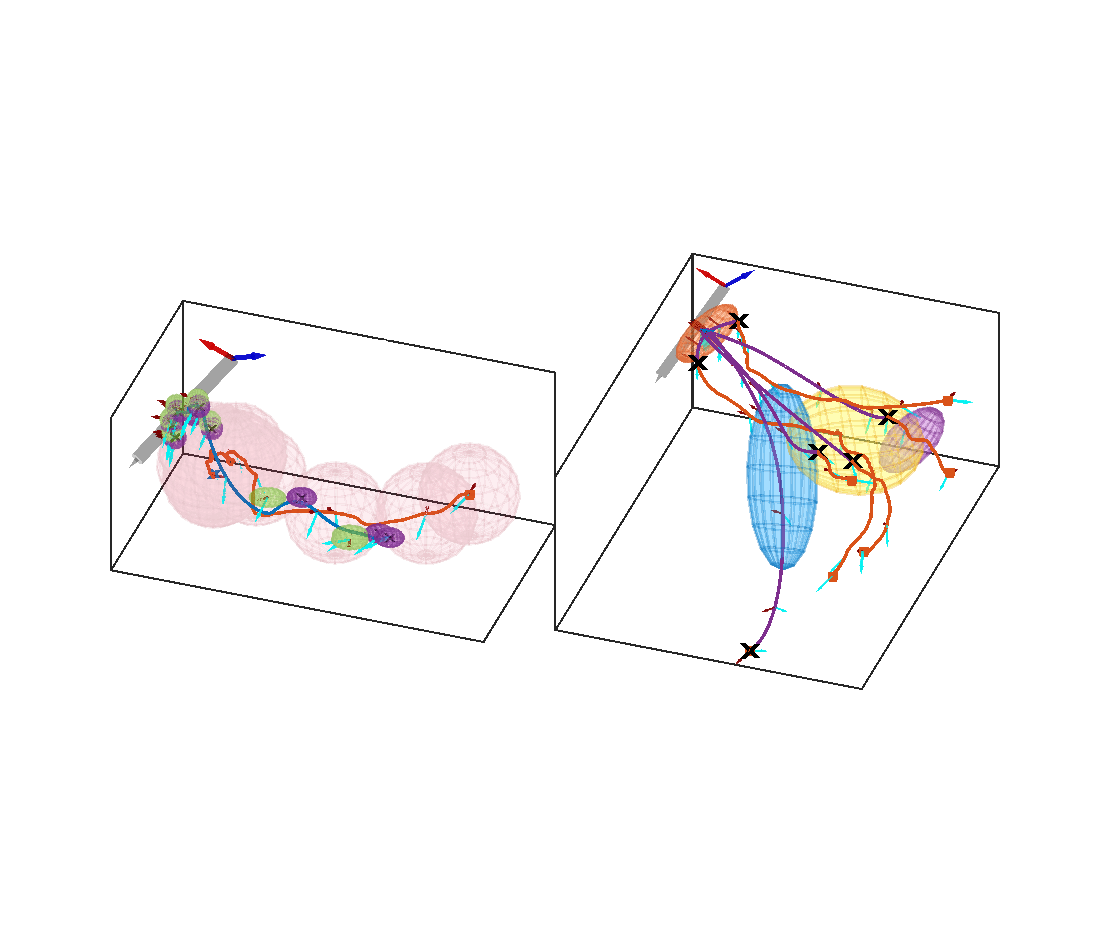} 
\end{center}
\caption{Semi-autonomous teleoperation for a new target pose with a screwdriver \textit{(top)} and a carabiner \textit{(bottom)}. Shared control example \textit{(left)}: the teleoperator demonstration (in red) strays away from the target pose, while the corrected trajectory (in blue) reaches the target pose. Desired state $\mathcal{N}(\mb{\hat{\mu}}_{t}, \mb{\hat{\Sigma}}_{t})$ is shown in purple, teleoperator state $\mathcal{N}(\mb{\xi}_t^{\ty{I}},\kappa^{2}\mb{I})$ in red, and predicted state $\mathcal{N}(\mb{\tilde{\mu}}_t^{\ty{O}}, \mb{\tilde{\Sigma}}_t^{\ty{O}})$ in green (see Sec.\ \ref{SecLP} for details). Autonomous control example \textit{(right)}: the arm movement is randomly switched (marked with a cross) from direct control (in red) to autonomous control (in purple) in which the learned model is used to generate the movement to the target pose.}
\label{fig: SCAC_SD}
\end{figure}

\begin{figure}[htp]
\begin{center}
\includegraphics[trim={2.5cm 0.4cm 2.0cm 0.75cm},clip,scale = 0.6]{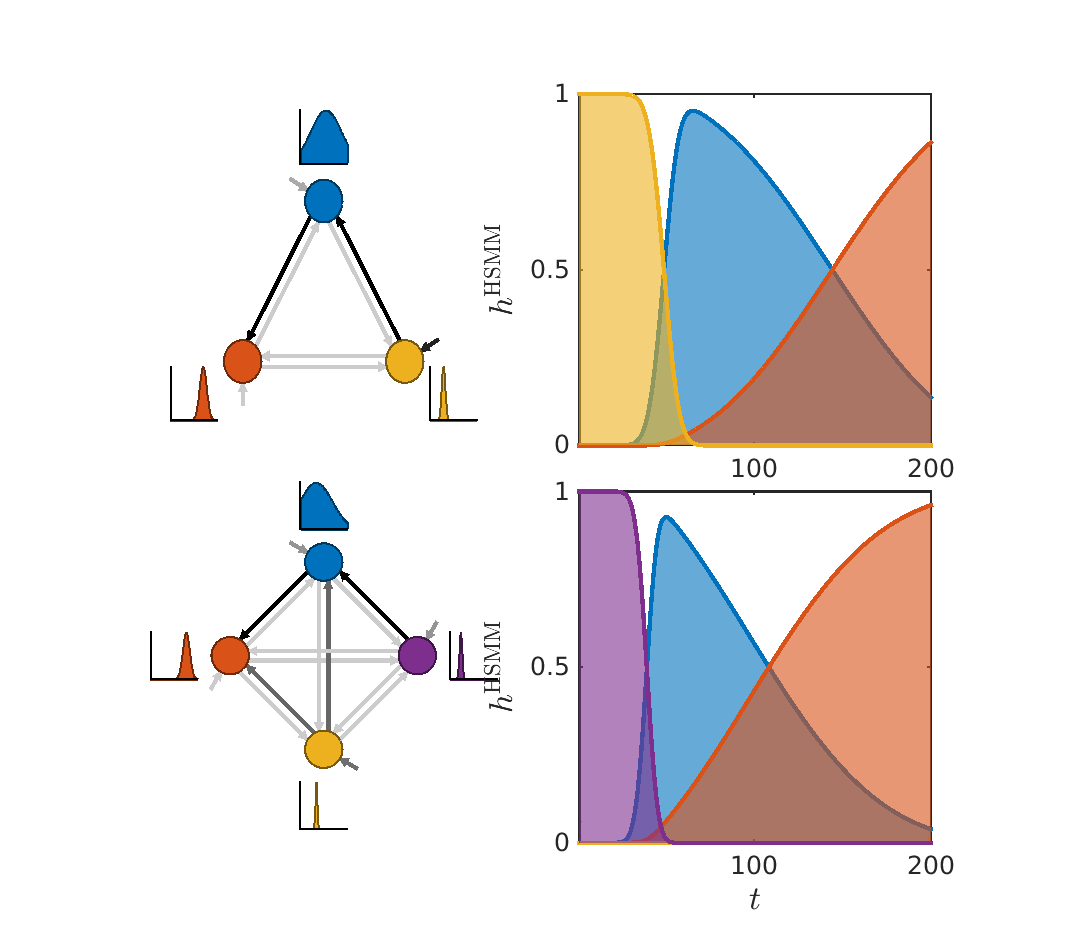}
\end{center}
\caption{HSMM graphical model representation ($s^{\max} = 150$) along with evolution of the rescaled forward variable for screwdriving (\textit{top}) and hooking a carabiner (\textit{bottom}).}\label{fig: SCAC_TM}
\end{figure}

\textbf{Time-Dependent Autonomous Control: } In case of communication disruptions, when it is difficult to retrieve the state of the teleoperator, the manipulation task can be completed in an autonomous manner. Task-parameterized SOSC model is used to generate the robot movement in an autonomous manner with the help of the forward variable $\alpha_{t,i}^{\tp{HSMM}} \triangleq P(z_t = i, \mb{\xi}_1 \ldots \mb{\xi}_t | \theta_{h})$. Given the model parameters $\theta_{h}$ and the partial observation sequence $\mb{\xi}_1 \ldots \mb{\xi}_t$, the probability of a datapoint $\mb{\xi}_t$ to be in state $i$ at time $t$ is recursively computed in the explicit duration HSMM using the forward variable as
\begin{multline}
	\label{Eq. HSMM_ForwardVariable}
	\alpha^{\tp{HSMM}}_{t,i} = \sum_{j=1}^{K} \sum_{s = 1}^{\min(s^{\max},t-1)}\alpha^{\tp{HSMM}}_{t-s,j}\; a_{j, i}\; 
	\mathcal{N}(s|\mu_{i}^{S},\Sigma_{i}^{S}) \\ 
	\prod_{c = t-s+1}^{t} \mathcal{N}(\mb{\xi}_c|\mb{\tilde{\mu}}_{i}, \mb{\tilde{\Sigma}}_{i}). 
\end{multline} 

The forward variable is used to evaluate the current state of the task $\mb{\xi}_{t_o}$ using $\alpha^{\tp{HSMM}}_{t_o,i} = \frac{\pi_i \mathcal{N}(\mb{\xi}_{t_o} |\mb{\tilde{\mu}}_i, \mb{\tilde{\Sigma}}_i)}{ \sum_{k=1}^{K}\pi_k \mathcal{N}(\mb{\xi}_{t_o} |\mb{\tilde{\mu}}_k, \mb{\tilde{\Sigma}}_k)}$, and subsequently plan the movement sequence for the next $T$ steps with $t = (t_o + 1) \ldots T$. Note that only the transition matrix and the duration model are used to plan the future evolution of the initial/current state $\mb{\xi}_{t_o}$ (the influence of the spatial data is omitted as it has not been observed), i.e., $\mathcal{N}(\mb{\xi}_{t}|\mb{\tilde{\mu}}_{i}, \mb{\tilde{\Sigma}}_{i}) = 1$ for $t = (t_o+1) \ldots T$. This is used to retrieve a stepwise reference trajectory $\mathcal{N}(\mb{\hat{\mu}}_t, \mb{\hat{\Sigma}}_t)$ from the state sequence $z_t$ computed from the forward variable. The stepwise reference trajectory is tracked in a smooth manner with a finite-horizon linear quadratic tracking controller \citep{Tanwani2016a} (see Appx.\ \ref{app: LQT_FH}), based on
\begin{equation}
	z_t = \argmax_i \; \alpha^{\tp{HSMM}}_{t,i}, \quad 
	\mb{\hat{\mu}}_t = \mb{\tilde{\mu}}_{z_t}^{\ty{O}}, \quad \mb{\hat{\Sigma}}_t = \mb{\tilde{\Sigma}}_{z_t}^{\ty{O}}.
\end{equation} 

We exploit the time-dependent autonomous control formulation to assist the teleoperator in performing challenging tasks and/or to counter large communication latencies. The teleoperator can switch at any time $t_o$ to the autonomous mode upon which the robot arm re-plans and executes the task for the next $T$ steps. When the task is accomplished or the communication channel is re-established, the operator can switch back to the manual control upon which the robot arm returns to the teleoperated state.
%

\begin{table}[tbp]
\small\centering
\caption{Performance comparison of the SOSC model against parametric batch HSMM models using number of parameters $N_p$, and the endpoint error between the teleoperated arm and the target. Teleoperation modes are direct control (DC), shared control (SC) and autonomous control (AC). Errors are in meters.} \normalsize \centering \label{tab_SOSC}
\begin{tabular}{|c||c|c|c|c|}
\hline
\multirow{2}{*}{\textbf{Model}} & \multirow{2}{*}{$N_{p}$} & \textbf{DC} & \textbf{SC} & \textbf{AC}\\
&  &   \textbf{Error} & \textbf{Error} & \textbf{Error}\\\hline \hline
\multicolumn{5}{|c|}{\textbf{screw-driving task ($K = 3, D = 14$)}}\\\hline
 \multirow{2}{*}{\textbf{FC-HSMM}}  & \multirow{2}{*}{$372$} &  & $0.095 \pm$ & $0.038 \pm$\\
& & & $0.025$ & $2.5 \times 10^{-5}$ \\\cline{1-2} \cline{4-5}
\multirow{2}{*}{\textbf{ST-HSMM}} & \multirow{2}{*}{$295$} &  & $0.094 \pm$ & $0.037 \pm$\\
 & & $0.30 \pm$ & $0.026$ & $1.8 \times 10^{-5}$ \\\cline{1-2} \cline{4-5}
\textbf{MFA-HSMM} & \multirow{2}{*}{$267$} & $0.17$ & $0.099 \pm$ & $\textbf{0.037} \pm$\\
\textbf{($d_k = 4$)} & &  & $0.022$ & $7.7 \times 10^{-6}$ \\\cline{1-2} \cline{4-5}
\textbf{\textbf{SOSC}} & \multirow{2}{*}{$\textbf{211}$} &  & $\textbf{0.084} \pm$ & $0.043 \pm$\\
\textbf{($\bar{d}_k = 3.67$)} &  & & $0.018$ & $1.3 \times 10^{-4}$ \\\hline\hline
\multicolumn{5}{|c|}{\textbf{hooking carabiner task ($K = 4, D = 14$)}}\\\hline
 \multirow{2}{*}{\textbf{FC-HSMM}}  & \multirow{2}{*}{$500$} &  & $0.081 \pm$ & $0.099 \pm$\\
& & & $0.056$ & $0.068$ \\\cline{1-2} \cline{4-5}
\multirow{2}{*}{\textbf{ST-HSMM}} & \multirow{2}{*}{$332$} &  & $0.082 \pm$ & $\textbf{0.022} \pm$\\
 & & $0.10 \pm$ & $0.058$ & $2.6 \times 10^{-4}$ \\\cline{1-2} \cline{4-5}
\textbf{MFA-HSMM} & \multirow{2}{*}{$360$} & $0.062$ & $\textbf{0.08} \pm$ & $0.037 \pm$\\
\textbf{($d_k = 4$)} & &  & $0.056$ & $8.8 \times 10^{-4}$ \\\cline{1-2} \cline{4-5}
\textbf{\textbf{SOSC}} & \multirow{2}{*}{$\textbf{318}$} &  & $\textbf{0.08} \pm$ & $0.073 \pm$\\
\textbf{($\bar{d}_k = 4.25$)} &  & & $0.056$ & $3.7 \times 10^{-4}$ \\\hline
\end{tabular}
\end{table}

\subsubsection{Results and Discussions:} 
We collect $6$ kinesthetic demonstrations for screwdriving with the initial pose of the target rotated/translated in the successive demonstrations, and perform $11$ demonstrations of hooking a carabiner at various places on the rod for $3$ different rotated configurations of the rod segment. Demonstrations are subsampled to $200$ datapoints for each demonstration, corresponding to an average of $7$ Hz. The parameters are defined as $\{\lambda =0.65, \lambda_1 = 0.03, \lambda_2 = 0.001, \lambda_3 = 0.04, \sigma^{2} = 2.5 \times 10^{-4}, \kappa^{2} = 0.01\}$.

Results of the task-parameterized SOSC model for the two tasks are shown in Fig.\ \ref{fig: JointDistributionSOSC}. We observe that the model exploits the variability in the demonstrations to statistically encode different phases of the task in the joint distribution. Demonstrations corresponding to the input component of the reference frame encode the reaching movement to different target poses with the screwdriver and the carabiner in the global frame, while the output component of the reference frame represents this movement observed from the viewpoint of the target (respectively shown as converging to a point for the screwdriver and to a line for the carabiner). The learned model for the screwdriving task contains $3$ clusters with subspace dimensions $\{4, 3, 4\}$, while the carabiner task model contains $4$ clusters with subspace dimensions $\{5, 5, 4 ,3\}$.

Fig.\ \ref{fig: SCAC_SD} (\emph{left}) shows how the model adjusts the movement of the teleoperator based on his/her current state in a time-independent manner. When the teleoperator is away from the target, the variance in the output distribution $\mathcal{N}(\mb{\tilde{\mu}}_t^{\ty{O}}, \mb{\tilde{\Sigma}}_t^{\ty{O}})$ is high and its product with the teleoperator frame $\mathcal{N}(\mb{\xi}_t^{\ty{I}},\kappa^{2}\mb{I})$ yields the desired state $\mathcal{N}(\mb{\hat{\mu}}_{t}, \mb{\hat{\Sigma}}_{t})$ closer to the teleoperator as in direct teleoperation. As the teleoperator moves closer to the target and visits low variance segments of $\mathcal{N}(\mb{\tilde{\mu}}_t^{\ty{O}}, \mb{\tilde{\Sigma}}_t^{\ty{O}})$, the desired state moves closer to the target as compared to the teleoperator. Consequently, the shared control formulation corrects the movement of the teleoperator when the teleoperator is straying from the target. Table \ref{tab_SOSC} shows the performance improvement of shared control over direct control where the endpoint error is reduced from $0.3$ to $0.084$ meters for the screwdriving task, and from $0.1$ to $0.08$ meters for the carabiner task. Error is measured at the end of the demonstration from the end-effector of the teleoperated arm to the target of the screwdriver, and to the rod segment for hooking the carabiner (see Appx.\ \ref{app: MnE}-$1$ for video of the semi-autonomous teleoperation experiments and results). 

To evaluate the autonomous control mode of the task-parameterized SOSC model, the teleoperator performs $6$ demonstrations and switches to the autonomous mode randomly while performing the task. The teleoperated arm evaluates the current state of the task and generates the desired sequence of states to be visited for the next $T$ steps using the forward variable, as shown in Fig.\ \ref{fig: SCAC_TM}. Fig.\ \ref{fig: SCAC_SD} (\emph{right}) shows that the movement of the robot converges to the target from different initial configurations of the teleoperator. As shown in Table \ref{tab_SOSC}, the obtained results are repeatable and more precise than the direct and the shared control results. Table \ref{tab_SOSC} also compares the performance of the SOSC algorithm against several parametric batch versions of HSMMs with different covariance models in the output state distribution, including full covariance (FC-HSMM), semi-tied covariance (ST-HSMM), and MFA decomposition of covariance (MFA-HSMM). Results of the SOSC model are used as a benchmark for model selection of the batch algorithms. We can see that the proposed non-parametric online learning model gives comparable performance to other parametric batch algorithms with a more parsimonious representation (reduced number of model parameters).

In our future work, we plan to bootstrap the online learning process with the batch algorithm after a few initial demonstrations of the task. We would like to use the initialized model to make a guess about the penalty parameters for non-parametric online learning. Moreover, we plan to test the model under more realistic environments with large communication latencies typically observed in satellite communication.


\section{Conclusions}\label{sec: Conclusion}

Non-parametric online learning is a promising way for adapting a model of movement behaviors while new training data are acquired. In this paper, we have presented a non-parametric scalable online sequence clustering algorithm by online inference in DP-MPPCA and HDP-HSMM under small variance asymptotics. The algorithm incrementally clusters the streaming data with non-parametric locally linear principal component analysis, and encodes the spatio-temporal patterns using an infinite hidden semi-Markov model. Non-parametric treatment gives the flexibility to continuously adapt the model with new incoming data. Learning the model online from a few human demonstrations is a pragmatic approach to teach new skills to robots. The proposed skill encoding scheme is potentially applicable to a wide range of tasks, while being robust to varying environmental conditions with the task-parameterized formulation. We show the efficacy of the approach to learn manipulation tasks online for semi-autonomous teleoperation, and assist the operator with shared control and/or autonomous control when performing remote manipulation tasks. 

%

%
%

\begin{funding}
This work was in part supported by the DexROV project through the EC Horizon 2020 programme (Grant \#635491).
\end{funding}



\bibliographystyle{SageH.bst}
\bibliography{0-bibliography-short}

%
%
%
\clearpage
\begin{appendices}
\section{Index to Multimedia Extensions} \label{app: MnE}
\begin{table}[!h]
\caption{Index to multimedia extensions.}
\label{tab:notation}
\begin{tabular}{|ccl|}
\hline & & \\
\textbf{Ext} & \textbf{Type} & \textbf{Description} \\[3mm] \hline\hline & & \\

$1$ & video & semi-autonomous teleoperation results \\[1mm]

$2$ & video & SOSC simulations  \\[1mm]
$3$ & codes & algorithms, experiments and datasets \\[1mm]\hline

\end{tabular}
\end{table}
\section{Symbols and Descriptions} \label{app: symbols}
%
%
%
%
%
%

\begin{table}[!h]
\caption{Description of symbols.}
\label{tab:notation}
\begin{tabular}{|cl|}
\hline & \\
\textbf{Symbol} & \textbf{Description}\\[3mm] \hline\hline &  \\

$\mb{\xi}_t$ & observation at time $t$ of dimension $D$\\[1mm]

$z_t $ & hidden state of $\mb{\xi}_t$ in $\{1 \ldots K \}$\\[1mm]

$\mb{a}_t$ & transition matrix with entries $a_{t, i,j}$\\[1mm]

$\{\mu_{t, i}^{S}, \Sigma_{t,i}^{S}\}$ & state duration mean and variance \\[1mm]

$\{\mb{\mu}_{t,i}, \mb{\Sigma}_{t,i}\}$  & output state distribution parameters\\[1mm]

$d_{t,i}$ & subspace dimension of state/cluster \\[1mm]

$\mb{\Lambda}_{t,i}^{d_{t,i}}$ & projection matrix of $d_{t,i}$ eigen vectors\\[1mm]

$\mb{U}_{t,i}^{d_{t,i}}$ & $d_{t,i}$ basis vectors of $\mb{\Sigma}_{t,i}$\\[1mm]

$b_m$ & bandwidth parameter to limit cluster size\\[1mm]

$w_{t,i}$ & weight of parameter set at time $t$\\[1mm]

$\mb{g}_{t,i}$ & projection of $\mb{\xi}_t$ on $\mb{U}_{t,i}^{d_{t,i}}$\\[1mm]

$\mb{p}_{t,i}$ & retro-projection of $\mb{g}_{t,i}$ in original space\\[1mm]

$\mb{R}_{t,i}$ & rotation matrix to update $\mb{U}_{t,i}^{d_{t,i}}$\\[1mm]

$\mb{\delta}_{i}$ & distance of $\mb{\xi}_t$ to each subspace of $\mb{U}_{t,i}^{k}$\\[1mm]

$\mb{\bar{e}}_{t,i}$ & average distance vector of $\mb{\delta}_i$\\[1mm]

$n_{t,i}$ & state transitions count from $i$ till time $t$ \\[1mm]

$\mb{c}_{t,i,j}$ & state transitions count from $i$ to $j$\\[1mm]

$s_{t}$ & duration steps count \\[1mm]

$\lambda$ & penalty for number of states \\[1mm]

$\lambda_1$ & penalty for subspace dimension \\[1mm]

$\lambda_2$ & penalty for transition to less visited states \\[1mm]

$\lambda_3$ & penalty for transition to unvisited state \\[1mm]\hline
\end{tabular}
\end{table}

\section{Infinite Horizon Linear Quadratic Regulator}\label{app: LQR_IH}
The desired reference state $\mathcal{N}(\mb{\hat{\mu}}_{t_0}, \mb{\hat{\Sigma}}_{t_0})$ can be smoothly followed by using an infinite-horizon linear quadratic regulator with a double integrator system. The cost function to minimize at current time step $t_0$ is given by
\begin{gather}
c(\mb{\xi}_t, \mb{u}_t) = \sum_{t=t_0}^{\infty}  (\mb{\xi}_t - \mb{\hat{\mu}}_{t_0})^{\trsp} \mb{Q}_{t_0} (\mb{\xi}_t - \mb{\hat{\mu}}_{{t_0}}) + \mb{u}_t^{\trsp} \mb{R} \mb{u}_t,  \nonumber \\
\mathrm{s.t.} \quad \mb{\dot{\xi}}_t = \begin{bmatrix}\mb{0} \quad \mb{I} \\ \mb{0} \quad \mb{0}\end{bmatrix} \mb{\xi}_t + \begin{bmatrix}\mb{0}\\ \mb{I}\end{bmatrix} \mb{u}_t, \nonumber
\end{gather} 
where $\mb{u}_t \in \mathbb{R}^{m}$ is the control input of the system. Setting $\mb{Q}_{t_0} = \mb{\hat{\Sigma}}_{t_0}^{-1}, \mb{R} \succ 0, \mb{\xi}_t = [{\mb{x}_t}^{\trsp} \; {\mb{\dot{x}}_t}^{\trsp}]^{\trsp}$, $\mb{\hat{\mu}}_{t_0} = [\mb{\hat{\mu}}_{t_0}^{x^{\trsp}}\; \mb{\hat{\mu}}_{t_0}^{\dot{x}^{\trsp}}]^{\trsp}$ with $\mb{x}, \mb{\dot{x}}$ representing the position and velocity of the system, the optimal control input $\mb{u}_t^{*}$ obtained by solving the algebraic Riccati equation is given by
\begin{equation*}
\mb{u}_t^{*} = \mb{K}^{\ty{P}}_t (\mb{\hat{\mu}}_{t_0}^{x} - \mb{x}_t) + \mb{K}^{\ty{V}}_t (\mb{\hat{\mu}}_{t_0}^{\dot{x}} - \mb{\dot{x}}_t),
\end{equation*} where $\mb{K}^{\ty{P}}_t$ and $\mb{K}^{\ty{V}}_t$ are 
the full stiffness and damping matrices for following the desired reference state.

\section{Finite Horizon Linear Quadratic Tracking} \label{app: LQT_FH}
Consider a double integrator system as an analogue of a unit mass attached to the datapoint $\mb{\xi}_t$. The desired step-wise reference trajectory $\mathcal{N}(\mb{\hat{\mu}}_t, \mb{\hat{\Sigma}}_t)$ is smoothly tracked by minimizing the cost function
\begin{gather}
c_t(\mb{\xi}_t, \mb{u}_t) = \sum_{t=1}^{T} (\mb{\xi}_t - \mb{\hat{\mu}}_t)^{\trsp} \mb{Q}_t (\mb{\xi}_t - \mb{\hat{\mu}}_t) + \mb{u}_t^{\trsp} \mb{R}_t \mb{u}_t,  \nonumber \\
\mathrm{s.t.} \quad \mb{\dot{\xi}}_t = \begin{bmatrix}\mb{0} \quad \mb{I} \\ \mb{0} \quad \mb{0}\end{bmatrix} \mb{\xi}_t + \begin{bmatrix}\mb{0}\\ \mb{I}\end{bmatrix} \mb{u}_t, \nonumber
\end{gather} starting from the initial state $\mb{\xi}_1$. Let $\mb{\xi}_t = [{\mb{x}_t}^{\trsp} \; {\mb{\dot{x}}_t}^{\trsp}]^{\trsp}, \mb{\hat{\mu}}_t = [{\mb{\hat{\mu}}_t^{x}}^{\trsp} \; {\mb{\hat{\mu}}_t^{\dot{x}}}^{\trsp}]^{\trsp}$ where $\mb{x}, \mb{\dot{x}}$ represent the position and velocity of the double integrator system. Setting $\mb{Q}_t = \mb{\hat{\Sigma}}_t^{-1} \succeq 0, \mb{R}_t \succ 0$, the control input $\mb{u}_t^{*}$ that minimizes the cost function is given by
\begin{eqnarray}
\mb{u}_t^{*} &=& - \mb{R}_t^{-1} \mb{B}_d^{\trsp} \mb{P}_t (\mb{\xi}_t - \mb{\hat{\mu}}_t) + \mb{R}_t^{-1} \mb{B}_d^{\trsp}\mb{d}_t,  \nonumber \\
& =& \mb{K}^{\ty{P}}_t (\mb{\hat{\mu}}_t^{x} - \mb{x}_t) + \mb{K}^{\ty{V}}_t (\mb{\hat{\mu}}_t^{\dot{x}} - \mb{\dot{x}}_t) + \mb{R}_t^{-1} \mb{B}_d^{\trsp}\mb{d}_t,\nonumber
\end{eqnarray} where $[\mb{K}^{\ty{P}}_t, \mb{K}^{\ty{V}}_t] = \mb{R}_t^{-1} \mb{B}_d^{\trsp} \mb{P}_t$ are the full stiffness and damping matrices, $\mb{R}_t^{-1} \mb{B}_d^{\trsp}\mb{d}_t$ is the feedforward term, and $\mb{P}_t, \mb{d}_t$ are the solutions of the differential equations
\begin{eqnarray}
-\mb{\dot{P}}_t & =& \mb{A}_d^{\trsp}\mb{P}_t + \mb{P_t}\mb{A}_d - \mb{P}_t\mb{B}_d\mb{R}_t^{-1}\mb{B}_d^{\trsp}\mb{P}_t + \mb{Q}_t,  \nonumber \\
-\mb{\dot{d}}_t &=& \mb{A}_d^{\trsp}\mb{d}_t - \mb{P}_t\mb{B}_d\mb{R}_t^{-1}\mb{B}_d^{\trsp}\mb{d}_t + \mb{P}_t\mb{\hat{\dot{\mu}}}_t - \mb{P}_t\mb{A}_d \mb{\hat{\mu}}_t, \nonumber
\end{eqnarray} with terminal conditions set to $\mb{P}_T = 0$ and $\mb{d}_T = 0$. Note that the gains can be precomputed before simulating the system if the reference trajectory and/or the task parameters do not change during the reproduction of the task. The resulting trajectory $\mb{\xi}_t^{*}$ smoothly tracks the stepwise reference trajectory $\mb{\hat{\mu}}_t$ and the gains $\mb{K}^{\ty{P}}_t, \mb{K}^{\ty{V}}_t$ stabilize $\mb{\xi}_t$ along $\mb{\xi}_t^{*}$ in accordance with the precision required during the task. 
\end{appendices}
\end{document}